\documentclass[sigconf]{acmart}  %
\AtBeginDocument{%
  }

\usepackage[dvipsnames]{xcolor}
\usepackage{soul}
\usepackage[breakable,skins]{tcolorbox}
\usepackage{adjustbox}
\usepackage[capitalise]{cleveref}
\Crefname{figure}{Fig.}{Figs.}
\Crefname{table}{Tab.}{Tabs.} 
\Crefname{section}{Sec.}{Secs.} 
\Crefname{subsection}{Sec.}{Secs.}
\Crefname{equation}{Eq.}{Eqs.}
\usepackage[T1]{fontenc}
\usepackage{newtxtext}
\usepackage{bm}
\usepackage{amsmath}
\DeclareMathAlphabet{\mymathbb}{U}{BOONDOX-ds}{m}{n}  %
\usepackage{xspace} 
\usepackage{multirow} 
\usepackage{tabularx}
\usepackage{marvosym}
\usepackage{colortbl}
\usepackage{pifont}
\usepackage[normalem]{ulem}
\usepackage{enumitem}
\usepackage{graphicx}
\graphicspath{{figures/}}
\usepackage{relsize}
\usepackage{nicematrix}
\usepackage{wrapfig}
\usepackage{makecell}
\usepackage{cleveref}
\usepackage{adjustbox}
\usepackage{subfig}
\usepackage{balance}
\usepackage{bbm}
\usepackage{comment}
\usepackage{enumitem}
\usepackage{xspace}

\definecolor{goodgreen}{rgb}{0.0, 0.56, 0.0}
\definecolor{badgray}{HTML}{666666}
\definecolor{highlight}{rgb}{0.98, 0.94, 1.0}
\definecolor{ourpurple}{RGB}{225, 213, 231}
\definecolor{ourred}{RGB}{234, 107, 102}
\definecolor{ouryellow}{RGB}{230, 195, 92}
\definecolor{ourorange}{RGB}{180, 101, 4}
\definecolor{ourgreen}{RGB}{213, 232, 212}
\definecolor{ourblue}{RGB}{108, 142, 191}
\definecolor{ourgray}{RGB}{102, 102, 102}

\let\titleold\title
\renewcommand{\title}[1]{\titleold{#1}\newcommand{\thetitle}{#1}}
\def\maketitlesupplementary
   {
   \newpage
       \twocolumn[
        \centering
        \Large
        \textbf{\thetitle}\\
        \vspace{0.5em}Supplementary Material \\
        \vspace{1.0em}
       ] %
   }

\usepackage{calc}

\newlength{\DepthReference}
\newlength{\HeightReference}
\newlength{\Width}%

\newcommand{\MyColorBox}[2][red]%
{%
    \settowidth{\Width}{#2}%
    \fcolorbox{#1}{white}%
    {%
        \raisebox{-\DepthReference}%
        {%
                \parbox[b][\HeightReference+\DepthReference][c]{\Width}{\centering\textcolor{#1}{#2}}%
        }%
    }%
}

\newcommand{\ie}{\textit{i.e.}\xspace}
\newcommand{\eg}{\textit{e.g.}\xspace}
\newcommand{\cmark}{\textcolor{ForestGreen}{\ding{51}}}
\newcommand{\xmark}{\textcolor{red}{\ding{55}}}
\newcommand{\method}{FineHOI\xspace}

\newcommand{\setobjects}{\mathbb{O}}
\newcommand{\setverbs}{\mathbb{V}}

\newcommand{\inputimage}{\mathbf{I}}

\newcommand{\setreal}{\mathbb{R}}
\newcommand{\adaptivepatchmodule}{Adaptive Part-Level Attention\xspace}
\newcommand{\objectdetector}{\mathcal{OD}}
\newcommand{\regiontransformermodule}{Region-Aware Interaction Transformer\xspace}
\newcommand{\supmat}{Supp. Mat.}

\definecolor{ourlightgray}{rgb}{0.83, 0.83, 0.83}
\definecolor{ourblue}{HTML}{89c2d9}
\definecolor{ourpink}{HTML}{AE4132}
\definecolor{ourorange}{HTML}{B46504}
\definecolor{ouryellow}{HTML}{D6B656}

\copyrightyear{2026}
\acmYear{2026}
\setcopyright{cc}
\setcctype{by-nc-nd}
\acmConference[MM '26]{Proceedings of the 34th ACM International Conference on Multimedia}{November 10--14, 2026}{Rio de Janeiro, Brazil}
\acmBooktitle{Proceedings of the 34th ACM International Conference on Multimedia (MM '26), November 10--14, 2026, Rio de Janeiro, Brazil}
\acmDOI{10.1145/3767308.3836016}
\acmISBN{979-8-4007-2213-4/2026/11}

\begin{document}

\title{FineHOI: Part-Aware Dense Representations for \\ Zero-Shot Human-Object Interaction Detection}  %

\author{Francesco Tonini}
\authornote{Work done while at the University of Amsterdam.}
\orcid{0000-0002-1938-3449}
\affiliation{%
  \institution{University of Trento}
  \city{Trento}
  \country{Italy}
}
\affiliation{%
  \institution{Fondazione Bruno Kessler}
  \city{Trento}
  \country{Italy}
}

\author{Lorenzo Vaquero}
\orcid{0000-0002-1874-3078}
\affiliation{%
  \institution{Fondazione Bruno Kessler}
  \city{Trento}
  \country{Italy}
}

\author{Mohammad Mahdi Derakhshani}
\orcid{0000-0003-0307-8439}
\affiliation{%
  \institution{University of Amsterdam}
  \city{Amsterdam}
  \country{The Netherlands}
}

\author{Cees Snoek}
\orcid{0000-0001-9092-1556}
\affiliation{%
  \institution{University of Amsterdam}
  \city{Amsterdam}
  \country{The Netherlands}
}

\author{Elisa Ricci}
\orcid{0000-0002-0228-1147}
\affiliation{%
  \institution{University of Trento}
  \city{Trento}
  \country{Italy}
}
\affiliation{%
  \institution{Fondazione Bruno Kessler}
  \city{Trento}
  \country{Italy}
}

\author{Cigdem Beyan}
\orcid{0000-0002-9583-0087}
\affiliation{%
  \department{Department of Computer Science}
  \institution{University of Verona}
  \city{Verona}
  \country{Italy}
}

\renewcommand{\shortauthors}{Francesco Tonini et al.}

\begin{abstract}
Human-Object Interaction (HOI) detection aims to localize humans and objects in images and classify their interactions.
Zero-shot HOI focuses on recognizing interactions that are not observed during training, requiring models to generalize beyond seen verb-object compositions. Recent approaches leverage Vision-Language Models (VLMs), benefiting from rich semantic representations. However, they often rely on global or detector-centric features that compress interaction cues and hinder fine-grained spatial reasoning.
To overcome this limitation, we propose \method, a zero-shot HOI framework that explicitly models interactions from dense patch-level features. Our approach is motivated by the observation that human-object interactions are defined by localized spatial relationships, which are not preserved by global and detector-centric representations.
To this end, we introduce an \adaptivepatchmodule module that decomposes humans and objects into semantically coherent parts via unsupervised clustering, and re-weights them based on their interaction relevance.
These representations are then integrated through a \regiontransformermodule that integrates part-aware and global features and produces the final HOI embedding.
Extensive experiments demonstrate that \method consistently outperforms existing zero-shot HOI methods, achieving particularly strong gains on unseen interactions.
Code is available \href{https://github.com/francescotonini/fine-hoi}{here}.
\end{abstract}



\maketitle

\section{Introduction}
\label{sec:main:intro}

\begin{figure}[!t]
    \centering
    \includegraphics[width=0.9\linewidth]{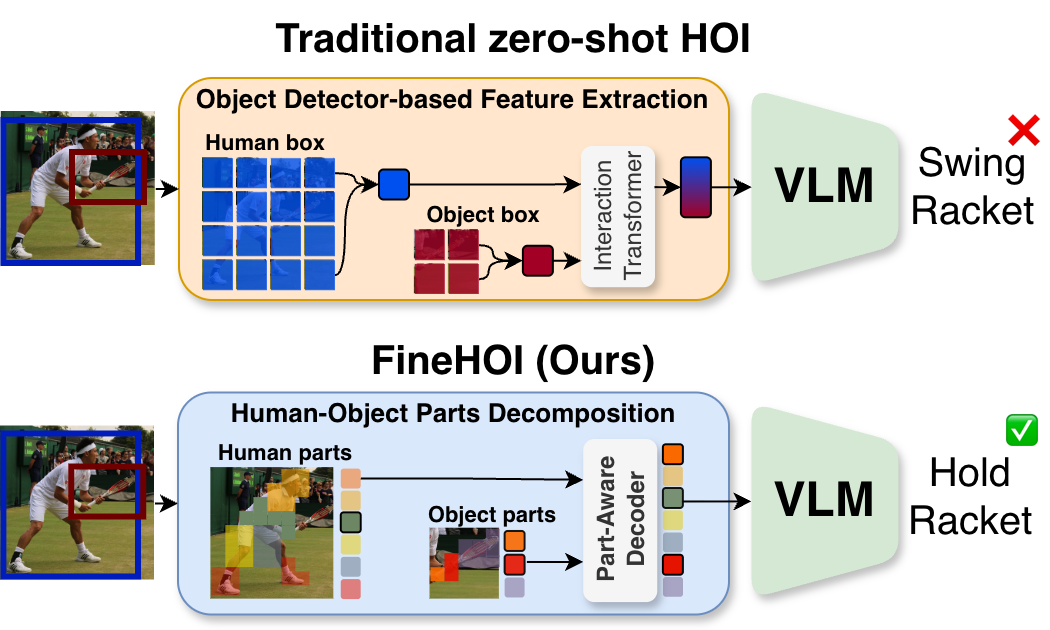}
    \caption{Comparison of zero-shot HOI approaches. Traditional methods, \eg \cite{kim2025LAIN,lei2024ez,lei2024exploring} rely on object-level features which do not capture the fine-grained details needed for interaction classification. In contrast, \method explicitly models dense patch-level interactions through human/object parts decomposition, and can dynamically select which parts are most relevant for an interaction (\eg hands, upper-body, object handle), enabling finer understanding of human-object interactions.
    }
    \label{fig:teaser}
    \vspace{-1.5em}
\end{figure}

Human-Object Interaction (HOI) detection aims to locate human-object pairs in images and identify their interactions as a set of $\langle$\textit{human}, \textit{verb}, \textit{object}$\rangle$ triplets.
HOI models are critical for a range of downstream applications, including image and video captioning~\cite{9583890,Seo_2022_CVPR,herdade2019image}, visual surveillance~\cite{li2024ripple}, and autonomous driving~\cite{chen2024asynchronous}, as they enhance perception and scene understanding in complex environments. Existing HOI methods~\cite{zhang2023exploring,Tu2022} rely heavily on large-scale annotated datasets~\cite{Gupta2015,Chao2018} to learn interaction representations. However, such annotations are expensive to obtain, inherently ambiguous, and difficult to scale to new domains or rare, long-tail interactions.

Recently, zero-shot HOI models have emerged as a promising direction to address these limitations~\cite{Bansal2020,VCL,lei2024ez,yang2025visual,kim2025LAIN,mao2023clip4hoi,lei2024exploring,guo2024unseen}. Unlike conventional closed-set HOI approaches~\cite{Kim2021,Tamura2021,Ulutan2020,zhang2021mining,zhang2023exploring,geng2025horp}, which are limited to recognizing interactions seen during training, zero-shot HOI models aim to generalize to unseen interactions. In this setting, models are trained on a subset of interaction categories and evaluated on disjoint unseen interactions, including novel verb-object compositions as well as unseen verbs or objects, requiring compositional generalization beyond observed data. This poses a significant challenge, as accurate recognition depends on capturing transferable and fine-grained interaction cues rather than memorizing combinations of verb-objects seen during training time.

This progress has been largely enabled by Vision-Language Models (VLMs), such as CLIP~\cite{radford2021learning}, which learn a shared multimodal embedding space that can be leveraged for zero-shot HOI detection.
This enables zero-shot recognition of novel interactions at test time by ranking text prompts for unseen verbs and objects against the visual embedding of the detected human-object pair.
Several recent works have explored this paradigm through techniques such as prompt learning~\cite{lei2024ez}, region-aware prompting~\cite{yang2025visual}, and lightweight adaptation of pretrained multimodal models~\cite{kim2025LAIN,mao2023clip4hoi,lei2024exploring}. Some approaches further incorporate auxiliary language supervision, derived from Large Language Models (LLMs)
~\cite{lei2024ez,yang2025visual}, while others focus on adapting VLMs without additional supervision~\cite{mao2023clip4hoi,kim2025LAIN,lei2024exploring}.

However, existing HOI detection methods rely on object detectors pretrained on large-scale datasets and build interaction representations from detector-centric features~\cite{mao2023clip4hoi,lei2024exploring,kim2025LAIN,yang2025visual}. As shown in~\cref{fig:teaser}, these features are typically instantiated as pooled region descriptors, object-centric embeddings, or detector queries, which summarize each human and object into a coarse representation. Although effective for localization and coarse semantic understanding, this early compression can discard the localized spatial cues needed to distinguish subtle interactions.
This early aggregation is particularly problematic in the zero-shot setting, where generalization depends on preserving transferable local evidence rather than relying on coarse object-level patterns. For example, as shown in~\cref{fig:teaser}, distinguishing whether a person is \textit{holding} a tennis racket rather than \textit{swinging} depends on localized cues around the hand-racket contact region together with the surrounding pose.
Closely related to our approach, LAIN~\cite{kim2025LAIN} introduces feature-level adaptations in the visual encoder to better capture locality. However, it still builds each human-object representation from detector-derived object-centric features, distilling the features into a single token for classification. 
Thus, although LAIN enhances the visual features, its final interaction representation remains compressed at the pair level, which can weaken the fine-grained features required to distinguish subtle HOIs.

To overcome these limitations, we propose \textbf{\method}, a zero-shot HOI framework that constructs interaction representations directly from dense image features instead of detector-centric embeddings.

We hypothesize that detector features, although effective for localization, are too object-centric and spatially coarse to support fine-grained HOI reasoning. Accordingly, we use the detector only to localize the regions of interest (\ie, human and object bounding boxes), and derive visual features from dense image patches rather than detector features.
Furthermore, motivated by the observation that HOI cues are often localized within specific contact regions (\eg, hands and racket handle, as in~\cref{fig:teaser}), rather than distributed across the entire instances, we introduce a novel \textit{\adaptivepatchmodule} module, which decomposes human and object regions into structurally coherent parts
and re-weights them based on their interaction relevance. This enables the model to focus on interaction-critical regions while preserving fine-grained spatial structure.

The resulting representations are processed by a novel \textit{\regiontransformermodule} that integrates interaction-relevant parts with aggregated patch features to form the final HOI representation, thereby preserving both the global context of the interaction and the localized evidence needed to disambiguate fine-grained visual relations in the zero-shot setting.
Experiments demonstrate consistent improvements across multiple zero-shot settings, with particularly strong gains in challenging scenarios involving unseen interactions. 

\noindent Our contributions are fourfold:

\begin{itemize} [leftmargin=*]
    \item
    We introduce \method, a zero-shot HOI detection method that builds interaction representations directly from dense patch-level features, avoiding early compression into object-level embeddings and preserving fine-grained spatial cues.

    \item 
    We propose an \adaptivepatchmodule module that performs unsupervised part decomposition with cross-entity attention to identify and emphasize interaction-relevant regions, enabling explicit modeling of localized human-object contact.

    \item We design a \regiontransformermodule that integrates part-aware representations with global context through dual-path reasoning, improving robustness and generalization to unseen interactions.

    \item
    We achieve consistent improvements across all zero-shot settings, outperforming prior methods while not relying on detector features or external LLM supervision.
\end{itemize}

\section{Related Work}
\label{sec:main:relatedwork}

\textbf{One-stage vs two-stage HOI Detection}. HOI detection aims to localize human-object pairs and classify their interactions. Existing methods can be broadly categorized into one-stage and two-stage approaches. One-stage methods formulate HOI detection as a set prediction problem, typically following DETR-like architectures with learnable queries to jointly predict humans, objects, and interactions. While effective in supervised closed-set settings, these methods require extensive annotations and do not generalize to unseen interactions~\cite{Kim2021,Kim2022,Tu2022,Zhong2022}.

Two-stage methods, on the other hand, first detect human and object instances using pretrained object detectors and then classify interactions for each candidate pair~\cite{Chao2018,Gupta2019,Gao2020,Zhang2021,Ulutan2020,zhang2022exploring}. These approaches leverage various cues, including spatial configurations~\cite{Gao2020,Zhang2021,Ulutan2020}, visual attention~\cite{Zhang2022}, and contextual reasoning~\cite{Liu2020b,Zhang2021,tonini2025dynamic,geng2025horp}, to model interactions. Due to their modular design and flexibility, two-stage pipelines are often adopted in settings requiring generalization, such as zero-shot HOI detection. Our method follows this paradigm. \\

\noindent \textbf{Zero-shot HOI Detection.}
Zero-shot HOI detection aims to recognize interactions that are not observed during training~\cite{Bansal2020}. Early approaches address this challenge through compositional learning~\cite{Bansal2020,VCL}, while more recent methods leverage VLMs, such as CLIP~\cite{radford2021learning}, to transfer semantic knowledge to unseen interactions~\cite{lei2024ez,kim2025LAIN,mao2023clip4hoi,lei2024exploring,cao2026semantic,guo2024unseen}. These approaches typically align visual features of human-object pairs with textual representations of interaction labels through prompt learning~\cite{lei2024ez}, proposal-based reasoning~\cite{yang2025visual}, or lightweight adaptation strategies~\cite{kim2025LAIN}.

\begin{figure*}
    \centering
    \includegraphics[width=0.9\linewidth]{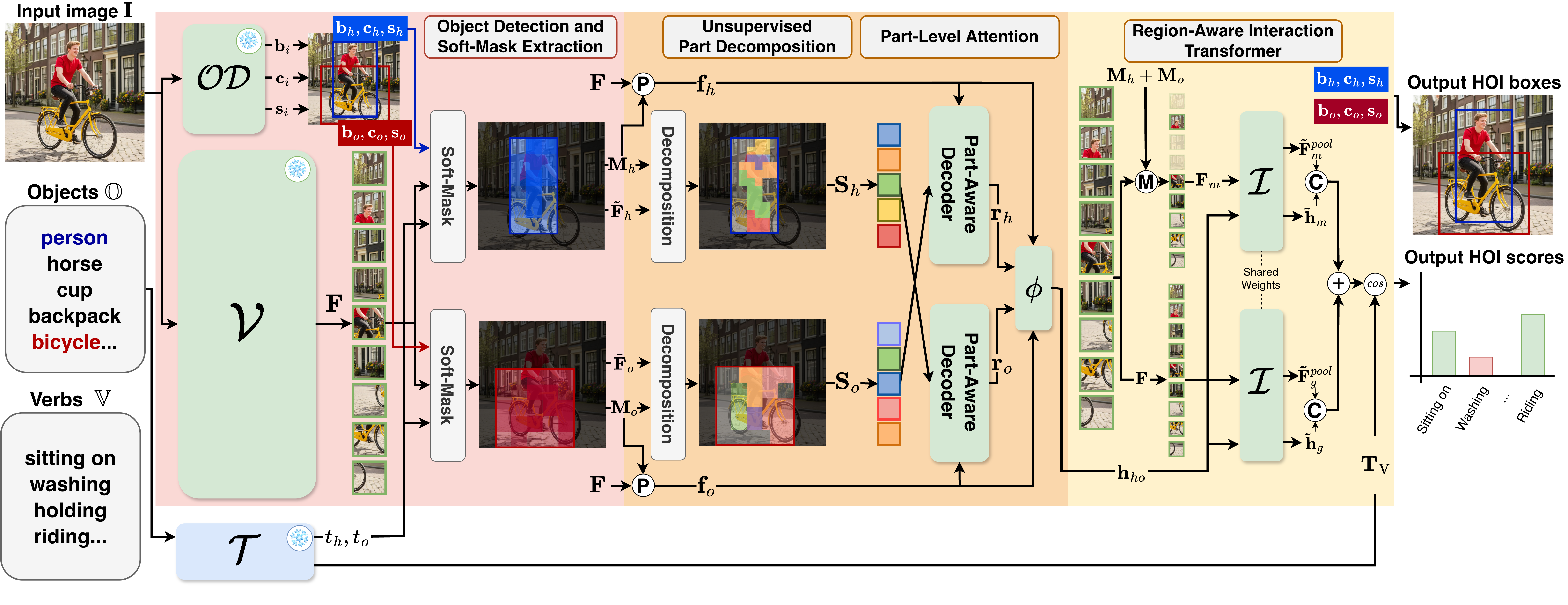}
    \vspace{-1em}
    \caption{
\method consists of three stages: \MyColorBox[ourpink]{Object Detection and Soft-Mask Extraction}, which uses an object detector $\mathcal{OD}$ to localize the human and object instances, while a visual encoder $\mathcal{V}$ extracts dense spatial feature map used for human and object soft masks construction; \protect\MyColorBox[ourorange]{\adaptivepatchmodule}, which decomposes the human and object patches into structurally coherent parts and refines them based on interaction relevance; and \MyColorBox[ouryellow]{Region-Aware Interaction Transformer}, which fuses the refined human-object representations with contextual patch features for HOI classification. \protect\textcircled{P} stands for feature pooling (\cref{eq:global_feature_pooling}), \protect\textcircled{M} for mask-guided feature pooling (\cref{eq:masked_feature_pooling}), and \protect\textcircled{C} for concatenation.}
\label{fig:main:method}
\vspace{-1em}
\end{figure*}

To improve generalization, recent works incorporate additional language information. Some approaches~\cite{lei2024ez,yang2025visual} exploit auxiliary language supervision to enrich interaction representations, while others~\cite{mao2023clip4hoi,kim2025LAIN,lei2024exploring} focus on adapting VLMs without additional supervision. However, all these methods rely on object detectors pretrained on large-scale datasets to generate human-object proposals~\cite{mao2023clip4hoi,lei2024exploring,kim2025LAIN,yang2025visual}, and build interaction representations from detector-centric features or global embeddings, which tend to summarize the human-object pair too early and can discard the localized visual cues needed to distinguish fine-grained interactions.

Despite these recent advances, these methods overlook a crucial aspect, \ie the explicit modeling of structured local spatial relationships, which is critical for fine-grained detection of HOIs. %
In contrast, our proposed \method explicitly models the interaction structure by decomposing image regions corresponding to the human and object into parts and reasoning over their relationships.
Rather than relying on global embeddings or detector-centric features, we construct interaction representations directly from dense patch-level visual features, enabling structured and localized HOI modeling.
Closely related to our approach, LAIN~\cite{kim2025LAIN} improves locality in the visual encoder. However, as shown in~\cref{fig:teaser}, it still relies on detector features and reduces each human-object pair to a single representation at an early stage. As a result, fine-grained contact cues are mostly lost.

\vspace{-0.5em}
\section{\method}
\label{sec:main:method}

\method builds interaction representations from the dense patch-level spatial features of a VLM visual encoder, enabling more fine-grained HOI recognition.
As illustrated in~\cref{fig:main:method}, given an input image, \method first employs an off-the-shelf object detector to localize human and object instances, then uses a VLM to extract dense patch-level spatial features and generate soft masks for the corresponding regions of interest (\cref{sec:main:method:object_detector_and_soft_mask}). 
Next, it partitions the human and object regions into structurally coherent parts and estimates their interaction relevance to produce refined part-aware representations (\cref{sec:main:method:adaptive_grounding}).
Finally, \method fuses the refined human-object representations into a joint interaction embedding, which is used for HOI classification.

\noindent \textbf{Problem Formulation.}
Given an input image $\inputimage \in \setreal^{H \times W \times 3}$ and candidate sets of object $\setobjects$ and verb $\setverbs$ labels, respectively, the goal of HOI detection is to locate and classify all interactions present in the image.
Formally, the $i$-th HOI instance is represented as $(\mathbf{b}_{h_{i}}, \mathbf{b}_{o_{i}}, o_{i}, v_{i})$, where $\mathbf{b}_{h_{i}}, \mathbf{b}_{o_{i}} \in \setreal^4$ denote the human and object bounding boxes, while $o_{i} \in \setobjects$ and $v_{i} \in \setverbs$ denote the object and verb label.
In the zero-shot HOI setting, training and evaluation are performed on disjoint sets of valid interaction classes, each defined by a $\langle \text{\textit{verb}}, \text{\textit{object}} \rangle$ combination. Specifically, the model is trained only on seen interactions and evaluated on unseen interactions that are not observed during training.

\subsection{Object Detection and Soft-Mask Extraction}
\label{sec:main:method:object_detector_and_soft_mask}

As seen in~\cref{fig:main:method} in red, an off-the-shelf object detector $\objectdetector$ processes the input image $\inputimage$ to produce a set of object proposals $\{d_i\}_{i=1}^N$, where each proposal
$d_i = (\mathbf{b}_i, s_i, c_i)$ consists of a bounding box
$\mathbf{b}_i \in \mathbb{R}^4$, a detection confidence score $s_i \in [0,1]$, and a class label $c_i \in \setobjects$. 
After non-maximum suppression, detections are partitioned into human and object boxes, and all possible human-object pairs are built.
For ease of notation, we consider a single human-object pair, where $d_h = (\mathbf{b}_h, s_h, c_h)$ and $d_o = (\mathbf{b}_o, s_o, c_o)$ denote the human and object detections, respectively.

Let $\mathcal{V}$ and $\mathcal{T}$ denote the visual and text encoders of a VLM, respectively, whose outputs lie in a shared embedding space.
Given $\inputimage$, the visual encoder $\mathcal{V}$ produces a spatial feature map $\mathbf{F} \in \mathbb{R}^{H_p \times W_p \times D}$, where $H_p \times W_p$ is the patch-grid resolution and $D$ is the embedding dimension.

For each patch-grid location $(x,y)$ of the image feature map $\mathbf{F}$, with $x \in \{1,\dots,H_p\}$ and $y \in \{1,\dots,W_p\}$, we define the human soft mask as: 
\begin{equation}
\mathbf{M}_h^{(x,y)} = \mathbbm{1}\!\left[(x,y) \in \mathbf{b}_h\right] s(\mathbf{F}^{(x,y)}, \mathbf{t}_h)
\end{equation}
where $s(\cdot,\cdot)$ denotes cosine similarity, $\mathbf{t}_h=\mathcal{T}(c_h) \in \mathbb{R}^{D}$ is the text embedding of the human, $c_h=\texttt{``person''}$, 
$\mathbbm{1}[\cdot]$ is the indicator function, and $(x,y) \in \mathbf{b}_h$ indicates that the patch-grid location $(x,y)$ lies within the human bounding box $\mathbf{b}_h$.
This formulation yields a mask $\mathbf{M}_h$ that suppresses patches outside the bounding box while preserving class-conditioned similarity within it. 
This enables the model to focus on semantically relevant regions while filtering out background noise, providing a more discriminative representation for interaction reasoning.
The object soft mask $\mathbf{M}_o$ is defined analogously using the object label $\mathbf{c}_o$ and the object bounding box $\mathbf{b}_o$.

\subsection{\adaptivepatchmodule}
\label{sec:main:method:adaptive_grounding}
While the soft masks from the previous stage indicate \emph{where} the human and the object are located within the proposed bounding boxes, they do not capture \emph{how} the two entities interact. 
HOIs are inherently local (\eg, a hand grasping a cup or a foot kicking a ball), and not all image regions associated with the human and the object contribute equally. 
Identifying the parts of the human and the object that are most relevant to the interaction is therefore crucial for extracting fine-grained features.

To address this, we introduce a novel \emph{\adaptivepatchmodule} module (\cref{fig:main:method}, orange block), which (a) partitions patch-level features into semantically-coherent parts, and then (b) uses said parts to adapt the corresponding visual features according to their relevance to the interaction.

\noindent \textbf{Unsupervised Part Decomposition.}
For each human and object, we first extract the patch features from $\mathbf{F}$ that lie within their respective bounding boxes, yielding the crop-level features $\tilde{\mathbf{F}}_h \in \mathbb{R}^{H_h \times W_h \times D}$ and $\tilde{\mathbf{F}}_o \in \mathbb{R}^{H_o \times W_o \times D}$, where $H_h \times W_h$ and $H_o \times W_o$ corresponds to the size of the human and object bounding boxes on the patch grid. 
By isolating these features before decomposition, we ensure that we focus exclusively on the human/object structure, avoiding interference from the surrounding context.

We then apply a modified version of Normalized Cut~\cite{shi2000normalized} (NCut) that acts on feature patches rather than pixels. This partitions $\tilde{\mathbf{F}}_h$ and $\tilde{\mathbf{F}}_o$ into $K$ parts, producing structurally coherent regions that capture meaningful object components, enabling more fine-grained interaction modeling than holistic region representations.
We derive the visual representation of each human part from its corresponding spatial region in the human soft mask $\mathbf{M}_h$.
Specifically, for each part $k \in \{1, \dots, K\}$, we obtain its features via mask-weighted pooling over the crop features:
\begin{equation}
\mathbf{s}_{h,k} =
\frac{\sum_{(x,y)\in \mathbf{P}_{h,k}} \mathbf{M}_h^{(x,y)}\,\tilde{\mathbf{F}}_h^{(x,y)}}
{\sum_{(x,y)\in \mathbf{P}_{h,k}} \mathbf{M}_h^{(x,y)}}.
\end{equation}
Here, $\mathbf{P}_{h,k}$ denotes the set of patch-grid locations belonging to the $k$-th human part.
The object part features $\mathbf{s}_{o,k}$ are defined analogously using $\mathbf{M}_o$, $\tilde{\mathbf{F}}_o$, and $\mathbf{P}_{o,k}$.
The use of $\mathbf{M}_h$, $\mathbf{M}_o$ ensures that background patches receive lower weight even within each part, keeping the representation focused on semantically relevant regions. 
This reduces noise from irrelevant context and improves the discriminability of interaction features.
We apply this procedure to all human and object parts, which yields the corresponding sets of part features,
$\mathbf{S}_h = \{ \mathbf{s}_{h,k} \}_{k=1}^{K}$ and $\mathbf{S}_o = \{ \mathbf{s}_{o,k} \}_{k=1}^{K}$,
for the human and object, respectively.

\noindent \textbf{Part-Aware Decoder.}
We argue that not all parts contribute equally to interaction prediction. 
For example, when predicting $\langle$\textit{human, riding, bike}$\rangle$, the parts corresponding to the hands and legs are more informative than those corresponding to the torso.
Accordingly, after decomposing humans and objects into parts, we use the cross-entity attention mechanism described below to adaptively modulate the contribution of each part.

First, we compute a pooled feature vector for the human branch by performing mask-weighted average pooling over all patch features:
\begin{equation}
\label{eq:global_feature_pooling}
\mathbf{f}_h = \frac{\sum_{x,y} \mathbf{M}_h^{(x,y)}\,\mathbf{F}^{(x,y)}}
{\sum_{x,y} \mathbf{M}_h^{(x,y)}}.
\end{equation}
The object feature vector $\mathbf{f}_o$ is defined analogously using the object soft mask $\mathbf{M}_o$.
These feature vectors will act as queries and attend to the part features of the other entity, enabling the model to adaptively identify interaction-relevant parts based on the pooled representation of each entity.
Specifically, the object feature vector $\mathbf{f}_o$ will attend over $\mathbf{S}_h$ to identify the interacting parts (\eg, feet), while the human feature vector $\mathbf{f}_h$ will attend over object parts $\mathbf{S}_o$ to identify the corresponding contact parts (\eg, pedals).

To this end, given the human feature vector $\mathbf{f}_h$, we compute an interaction-aware representation $\mathbf{r}_h \in \mathbb{R}^{d_v}$ via cross-attention:
\begin{equation}
\mathbf{Q} = \mathbf{f}_h \mathbf{W}_q,
\qquad
\mathbf{K} = \mathbf{S}_o \mathbf{W}_k,
\qquad
\mathbf{V} = \mathbf{S}_o \mathbf{W}_v,
\end{equation}
\begin{equation}
\mathbf{r}_h = \mathrm{Softmax}\!\left(\frac{\mathbf{Q}\mathbf{K}^\top}{\sqrt{d_v}}\right)\mathbf{V},
\end{equation}
where $\mathbf{W}_q \in \mathbb{R}^{D \times D}$, $\mathbf{W}_k \in \mathbb{R}^{D \times D}$, and $\mathbf{W}_v \in \mathbb{R}^{D \times D}$ are learned projection matrices. 
This formulation allows the model to attend to interaction-relevant parts.  By conditioning each entity on the parts of the other, it explicitly models the contact regions that define the interaction.
We also apply the above process to the object feature vector $\mathbf{f}_o$ using the human part features $\mathbf{S}_h$, yielding $\mathbf{r}_o$.

Finally, we combine the pooled and interaction-aware representations to obtain a unified human-object representation:
\begin{equation}
\mathbf{h}_{ho}
=
\phi\!\left(\mathbf{f}_h \,\|\, \mathbf{f}_o \,\|\, \mathbf{r}_h \,\|\, \mathbf{r}_o\right),
\end{equation}
where $\phi : \mathbb{R}^{4D} \rightarrow \mathbb{R}^{D}$ is a learned linear projection, and $\|$ denotes concatenation.
This 
vector combines global semantic information with localized interaction-aware cues from the attended parts.

\subsection{\regiontransformermodule}
\label{sec:main:method:interaction_transformer}
Each human-object token $\mathbf{h}_{ho}$ captures pair-specific interaction cues.
Nevertheless, a reliable prediction of the HOI also requires contextual information from the surrounding image.
To integrate these complementary sources of evidence, we introduce a dual-path architecture that explicitly models global and local information separately (\cref{fig:main:method}, yellow block).
It leverages a region-aware interaction transformer $\mathcal{I}$: an encoder-only Vision Transformer~\cite{dosovitskiy2020image} model with shared parameters but different visual inputs, combining scene-level context with interaction-relevant parts.

\noindent \textbf{In the global path,}
the human-object pair token $\mathbf{h}_{ho}$ is processed together with the image patch features $\mathbf{F}$ through the interaction transformer $\mathcal{I}$:
\begin{equation}
\tilde{\mathbf{h}}_{g}, \tilde{\mathbf{F}}_{g} = \mathcal{I}(\mathbf{h}_{ho} \,\|\ \mathbf{F}),
\end{equation}
where $\tilde{\mathbf{h}}_{g} \in \mathbb{R}^{D}$ denotes the updated pair token and $\tilde{\mathbf{F}}_{g} \in \mathbb{R}^{H_p \times W_p \times D}$ denotes the refined patch features.
We then construct the global HOI embedding $\mathbf{e}_{g} \in \mathbb{R}^{2D}$ by concatenating the updated pair token with the spatial average of the refined patch features ${\mathbf{F}^{{pool}}_{g}}$:
\begin{equation}
\mathbf{e}_{g}
=
\tilde{\mathbf{h}}_{g}
\,\|\,
\underbrace{
\frac{1}{H_p W_p}
\sum_{x=1}^{H_p}\sum_{y=1}^{W_p}
\tilde{\mathbf{F}}_{g}^{(x,y)}
}_{\mathbf{\hat{F}}^{pool}_{g}}.
\end{equation}

\noindent \textbf{In the mask-guided path,} the human-object pair token $\mathbf{h}_{ho}$ is jointly encoded with mask-guided patch features through the same interaction transformer $\mathcal{I}$.
We first define the mask-guided patch features as:
\begin{equation}
\label{eq:masked_feature_pooling}
\mathbf{F}_{m}^{(x,y)}
=
\left(\mathbf{M}_{h}^{(x,y)} + \mathbf{M}_{o}^{(x,y)}\right)\mathbf{F}^{(x,y)} ,
\end{equation}
and then apply the interaction transformer to the pair token and the mask-guided patch features:
\begin{equation}
\tilde{\mathbf{h}}_{m}, \tilde{\mathbf{F}}_{m} = \mathcal{I}(\mathbf{h}_{ho} \,\|\ \mathbf{F}_{m}),
\end{equation}
where $\tilde{\mathbf{h}}_{m} \in \mathbb{R}^{D}$ denotes the updated pair token and $\tilde{\mathbf{F}}_{m} \in \mathbb{R}^{H_p \times W_p \times D}$ represents the refined mask-guided patch features.
We finally construct the mask-guided HOI embedding $\mathbf{e}_{m} \in \mathbb{R}^{2D}$ combining both the pair token and the average masked feature patches ${\mathbf{F}^{{pool}}_{m}}$, to emphasize localized interaction cues within the human-object pair:
\begin{equation}
\mathbf{e}_{m} =
\tilde{\mathbf{h}}_{m} \,\|\, 
\underbrace{
\frac{ \sum_{x=1}^{H_p}\sum_{y=1}^{W_p}
\left(\mathbf{M}_{h}^{(x,y)} + \mathbf{M}_{o}^{(x,y)}\right)\tilde{\mathbf{F}}_{m}^{(x,y)} }
{ \sum_{x=1}^{H_p}\sum_{y=1}^{W_p}
\left(\mathbf{M}_{h}^{(x,y)} + \mathbf{M}_{o}^{(x,y)}\right) }
}_{\mathbf{\tilde{F}}^{{pool}}_{m}} .
\end{equation}

The two paths capture complementary cues: the global path preserves scene-level context, whereas the mask-guided path emphasizes features associated with high human-object mask responses.
Lastly, we $\ell_2$-normalize the two embeddings and combine them into the final HOI representation:
\begin{equation}
\label{eq:main:hoi_logits}
\mathbf{e}_{hoi} = \alpha \bar{\mathbf{e}}_{g} + \bar{\mathbf{e}}_{m},
\end{equation}
where $\alpha$ is a learnable scalar that balances the contribution of the global path.

\subsection{Training and Inference}
Following prior zero-shot HOI methods~\cite{kim2025LAIN,lei2024ez,lei2024exploring}, we cast HOI as a multi-label classification task. Let $\setverbs$ denote the set of verb classes, and let $\mathbf{T}_V \in \mathbb{R}^{|\setverbs| \times 2D}$ be the corresponding text embeddings produced by text encoder $\mathcal{T}$. We compute scores for each verb by comparing the HOI embedding $\mathbf{e}_{hoi}$ with all verb embeddings in the shared space:
\begin{equation}
\text{HOI}_\text{pred} = \sigma\left(s(\mathbf{e}_{hoi}, \mathbf{T}_V)/\tau\right),
\end{equation}
where $s(\cdot,\cdot)$ denotes cosine similarity, $\sigma$ is the sigmoid function, and $\tau$ is a learnable temperature.

\noindent \textbf{During training},
for each candidate pair, we construct a binary label vector indicating the valid verb classes and optimize the predicted scores with binary cross-entropy:
\begin{equation}
\mathcal{L}_{\text{hoi}} = \mathrm{FL}(\text{HOI}_\text{pred}, \text{HOI}_\text{gt}),
\end{equation}
where $\text{HOI}_\text{gt} \in \{0,1\}^{N_{\text{pair}} \times |\setverbs|}$ contains the target verb labels for all $N_\text{pair}$ human-object pairs, and FL is the binary focal loss~\cite{lin2017focal}.

\noindent \textbf{At inference},
following prior methods~\cite{kim2025LAIN,lei2024ez,mao2023clip4hoi} we define the HOI scores as:
\begin{equation}
\label{eq:main:inference}
\text{HOI}_\text{test} = \text{HOI}_\text{pred} \cdot \mathbf{s}_h^{\lambda} \cdot \mathbf{s}_o^{\lambda},
\end{equation}
where $\mathbf{s}_h, \mathbf{s}_o$ are the detection scores of the paired human and object instances, and $\lambda$ is a fixed hyper-parameter.

\section{Experiments}
\label{sec:main_experiments}

\noindent \textbf{Datasets.}
We evaluate \method on standard HOI detection benchmarks, HICO-DET~\cite{Chao2018} and V-COCO~\cite{Gupta2015}. HICO-DET contains 47,776 images, with 38,118 for training and 9,658 for testing, covering 600 HOI categories defined over 117 verbs and 80 objects. 
V-COCO is built on the COCO~\cite{lin2014microsoft} dataset and consists of 10,346 images, split into 2,533 training, 2,867 validation, and 4,946 test images, with annotations spanning 29 action categories across 80 object classes.

\noindent \textbf{Evaluation Metrics.}
Following the standard evaluation protocols used by state-of-the-art zero-shot HOI methods~\cite{kim2025LAIN,cao2026semantic,lei2024ez,yang2025visual,guo2024unseen}, we evaluate \method using mean average precision (mAP) under five zero-shot settings~\cite{hou2021detecting,guo2024unseen}: Unseen Verb (UV), where the 117 verbs are split into 97 seen and 20 unseen classes; Unseen Object (UO), where the 80 objects are split into 68 seen and 12 unseen classes; and three interaction-level settings, Unseen Combination Rare-First (UC-RF), Unseen Combination Non-Rare-First (UC-NF), and Unseen Combination (UC), where the 600 interactions are split into 480 seen and 120 unseen pairs, with the unseen set consisting of either the rarest interactions (RF), the most frequent interactions (NF), or a predefined subset, respectively.

We report performance on Seen and Unseen HOI classes, along with the overall average (Full) and harmonic mean (HM), which provides a balanced measure of generalization across seen and unseen categories.
Furthermore, we also evaluate against zero-shot HOI methods in the closed-set setting, and report performance on the Full set, as well as on the Rare and Non-Rare splits, where the 600 HOI categories are divided into 138 Rare classes with fewer than 10 training instances and 462 Non-Rare classes~\cite{Chao2018}. 
In V-COCO, we report average precision under the $AP^{S2}$ protocol
which evaluates only interactions involving an explicit semantic role object, such as $\langle$\textit{human}, \textit{holding}, \textit{phone}$\rangle$ or $\langle$\textit{human}, \textit{sitting on}, \textit{horse}$\rangle$~\cite{Gupta2015}.

\begin{table*}[!t]
\caption{State-of-the-art comparison of zero-shot HOI detection methods on the HICO-DET~\cite{Chao2018} dataset. Results for prior methods are taken from~\cite{kim2025LAIN,yang2025visual}, while results marked with $\dagger$ are reproduced using the official code. $\Delta$ indicates the use of detector-centric features. Best results are highlighted in bold, and second-best results are \underline{underlined}.}

\label{tab:main:hicodet_zs}
\resizebox{0.9\textwidth}{!}{%
\begin{tabular}{@{}l@{\hskip 4pt}lc|c@{\hskip 5pt}c@{\hskip 2pt}c@{\hskip 2pt}c|c@{\hskip 5pt}c@{\hskip 2pt}c@{\hskip 2pt}c|c@{\hskip 5pt}c@{\hskip 2pt}c@{\hskip 2pt}c|c@{\hskip 5pt}c@{\hskip 2pt}c@{\hskip 2pt}c|c@{\hskip 5pt}c@{\hskip 2pt}c@{\hskip 2pt}c@{}}
\toprule
\multicolumn{3}{c|}{} & \multicolumn{4}{c|}{\textbf{UV}}                                                                                                                                                                                    & \multicolumn{4}{c|}{\textbf{UO}}                                                                                                                                                                                                           & \multicolumn{4}{c|}{\textbf{UC-RF}}                                                                                                                                                                                                           & \multicolumn{4}{c|}{\textbf{UC-NF}} & \multicolumn{4}{c}{\textbf{UC}}                                                                                                                                                                                                           \\
\multicolumn{2}{c}{\multirow{-2}{*}{\textbf{Model}}} & \multicolumn{1}{c|}{\multirow{-2}{*}{$\Delta$}} & \textbf{Full} & \textbf{Seen} & \textbf{Unseen} & \textbf{HM} & \textbf{Full} & \textbf{Seen} & \textbf{Unseen} & \textbf{HM} & \textbf{Full} & \textbf{Seen} & \textbf{Unseen} & \textbf{HM} & \textbf{Full} & \textbf{Seen} & \textbf{Unseen} & \textbf{HM} & \textbf{Full} & \textbf{Seen} & \textbf{Unseen} & \textbf{HM} \\
\midrule
\rowcolor{ourblue} \multicolumn{23}{c}{\textbf{w/ LLM supervision}} \\
\midrule
ADA-CM~\cite{Lei2023}  & CLIP ViT-B/16  & \checkmark &  - & - & - & - & - & - & - & - & 33.01 & 34.35 & 27.36 & 30.86 & 31.39 & 31.13 & 32.41 & 31.77 & - & - & - & - \\
EZ-HOI~\cite{lei2024ez}  & CLIP ViT-L/14  & \checkmark &  32.32 & 33.49 & 25.10 & 29.30 & 32.27 & 32.06 & 33.28 & 32.67 & 33.13 & 34.15 & 29.02 & 31.59 & 31.17 & 30.55 & 33.66 & 32.11 & - & - & - & - \\
DYSCO~\cite{tonini2025dynamic}  & CLIP ViT-L/14  & \checkmark &  - & - & - & - & - & - & - & - & 25.24 & 23.96 & 30.36 & 27.16 & 25.18 & 24.58 & 27.56 & 26.07 & - & - & - & - \\
{VDRP}~\cite{yang2025visual}                                               & {CLIP ViT-L/14}  & \checkmark & {37.18} & \underline{38.16} & {31.16} & {34.31}                           
& \textbf{{37.81}} & \textbf{{37.50}} & {39.39} & \underline{38.41}
& \underline{38.13} & {38.48} & \underline{36.72} & \underline{37.58}                           
& \underline{36.46} & \underline{36.21} & {37.48} & \underline{36.83} & - & - & - & -                           \\
\midrule
\rowcolor{ourblue} \multicolumn{23}{c}{\textbf{w/o LLM supervision}} \\
\midrule
CLIP4HOI~\cite{mao2023clip4hoi} & CLIP ViT-B/16  & \checkmark & 26.02 & 31.14 & 30.42 & 30.78 & 27.71 & 33.25 & 32.11 & 32.68 & 28.47 & 35.48 & 34.08 & 34.78 & 31.44 & 28.26 & 28.90 & 28.58 & 32.11 & 33.25 & 27.71 & 30.48 \\
HOIGen~\cite{guo2024unseen} & CLIP ViT-B/16 & \checkmark & 32.34 & 34.31 & 20.27 & 33.32 & 33.48 & 32.90 & 36.35 & 32.63 & 33.86 & 34.57 & 31.01 & 32.79 & 33.08 & 32.86 & 33.98 & 33.42 & 33.44 & 34.23 & 30.26 & 32.25 \\
SSGR~\cite{cao2026semantic} & CLIP ViT-B/16  & \checkmark & 34.37 & 36.60 & 20.67 & 28.49 & 35.41 & 35.14 & 36.73 & 35.93 & 35.71 & 36.98 & 30.60 & 33.79 & 33.80 & 33.52 & 34.89 & 34.20 & - & - & - & - \\
CMMP~\cite{lei2024exploring} & CLIP ViT-L/14  & \checkmark & 36.38 & 37.28 & 30.84 & 33.75 & 36.74 & 36.15 & 39.67 & 37.83 & 37.13 & 37.42 & 35.98 & 36.69 & 35.13 & 33.53 & 33.52 & 34.50 & 36.56 & 37.15 & \underline{34.46} & \underline{35.75} \\
{LAIN}~\cite{kim2025LAIN} & {CLIP ViT-L/14}                  & {\checkmark} & \underline{37.20}      & {38.04}                             & \underline{32.05}                               & \underline{35.05}                           & \underline{37.60}                             & \underline{36.96}                             & \textbf{{40.78}}                               & \textbf{{38.87}}                           & \underline{38.13}                             & \underline{38.54}                             & {36.57}                               & {37.55}                          & {36.22}                             & {35.90}                             & \underline{37.52}                               & {36.71} & \underline{36.81} & \underline{37.95} & {32.25} & {35.10}                           \\
{LAIN~\cite{kim2025LAIN}}$^\dagger$ & DINOtxt ViT-L/16               & \checkmark & {35.43}      & {36.09}                             & {31.34}                               & {33.72}                           & {36.01}                             & {34.72}                             & {38.11}                               & {36.41}                                                      & {36.82}                             & {37.55}                             & {33.90}                               & {35.82}  & {33.89}                             & {34.09}                             & {33.09}                               & {33.59}  & 32.97 &	33.53	& 30.71 &	32.12                       \\
{LAIN~\cite{kim2025LAIN}}$^\dagger$ & DINOtxt ViT-L/16 & &                {34.50}      & {34.98}                             & {31.57}                               & {33.27}                           & {34.85}                             & {34.38}                             & {37.18}                               & {35.78}                           & {34.79}                             & {35.56}                             & {31.70}                               & {33.63}                           & {34.80}                             & {34.60}                             & {35.58}                               & {35.09}  & 34.80	& 35.11 &	33.59 &	34.35                         \\
\midrule
\textbf{\method} & DINOtxt ViT-L/16  &                                       & \textbf{38.10}      & \textbf{38.51}                             & \textbf{35.58}                               & \textbf{37.04} &                           {36.62} & {35.92} & \underline{40.13} & {38.03}                             & \textbf{38.39}                             & \textbf{38.55}                             & \textbf{37.79}                               & \textbf{38.17}                           & \textbf{36.58}                             & \textbf{36.30}                             & \textbf{38.11}                               & \textbf{37.15}     & \textbf{36.88}	& \textbf{38.00} &	\textbf{34.79}	& \textbf{36.10}                      \\ 
\bottomrule
\end{tabular}%
}
\end{table*}

\subsection{Implementation Details}
Following established practice in zero-shot HOI detection~\cite{lei2024ez,cao2026semantic,kim2025LAIN}, \method employs a frozen object detector, DETR~\cite{Carion2020}, to identify human and object instances. 
However, unlike state-of-the-art zero-shot HOI methods, we use it only as a region proposal network and do not rely on its features at any stage of the model.
We discard detections with confidence scores below $0.2$ and retain up to 15 human and object instances per image. 
Unless otherwise specified, we instantiate the vision $\mathcal{V}$ and text backbones $\mathcal{T}$ using DINOtxt~\cite{simeoni2025dinov3}. Consistent with recent zero-shot HOI methods~\cite{lei2024ez,kim2025LAIN}, all backbone parameters are kept frozen and adapted only through residual adapters. For unsupervised part decomposition, we adopt the NCut implementation of~\cite{yang2024alignedcutvisualconceptsdiscovery} and set the maximum number of parts to $K=10$. 
We train \method for 20 epochs using the AdamW optimizer~\cite{loshchilov2017decoupled} with a batch size of 4 on a single RTX A6000 GPU. The initial learning rate is set to $10^{-3}$ and is reduced to $10^{-4}$ after epoch 10.
At inference time, we set $\lambda = 2.8$ in~\cref{eq:main:inference} as in prior works~\cite{lei2024ez,cao2026semantic,kim2025LAIN}.
Additional implementation details are provided in the \supmat.

\subsection{Comparison with the State-of-the-art}

\begin{table}[!t]
\centering
\caption{Evaluation of zero-shot methods on closed-set HOI detection on HICO-DET~\cite{Chao2018} and V-COCO~\cite{Gupta2015} datasets. Results for prior methods are taken from~\cite{kim2025LAIN,yang2025visual}, while results marked with $\dagger$ are reproduced using the official code. $\Delta$ indicates the use of detector-centric features. Best results are in bold, second-best are \underline{underlined}.}
\label{tab:main_hicodet_closeset}
\resizebox{0.9\columnwidth}{!}{%
\begin{tabular}{@{}l@{\hskip 4pt}lc|c@{\hskip 8pt}c@{\hskip 3pt}c@{\hskip 3pt}c|c@{}}
\toprule
\multicolumn{2}{c}{\multirow{2}{*}{\textbf{Model}}} & \multicolumn{1}{c|}{\multirow{2}{*}{$\Delta$}} & \multicolumn{4}{c|}{\textbf{HICO-DET}}                                       & \textbf{V-COCO}        \\
\multicolumn{3}{c|}{}                        &  \textbf{Full}           & \textbf{Rare}           & \textbf{Non-rare}       & \textbf{HM}             & 
$\text{\textbf{AP}}^\text{\textbf{S2}}$            \\
\midrule
\rowcolor{ourblue} \multicolumn{8}{c}{\textbf{w/ LLM supervision}} \\
\midrule
ADA-CM~\cite{lei2024exploring} & CLIP ViT-L/14 & \checkmark & 38.40 & 37.52 & 38.66 & 38.09 & 63.9 \\ 
EZ-HOI~\cite{lei2024ez}                    & CLIP ViT-L/14 & \checkmark & 38.61          & 37.70          & 38.89          & 38.30          & 66.1          \\
DYSCO~\cite{tonini2025dynamic} & CLIP ViT-L/14 & \checkmark & 28.24 & 34.22 & 26.46 & 30.34 & 47.8 \\
VDRP~\cite{yang2025visual}$^\dagger$                  & CLIP ViT-L/14 & \checkmark  & \underline{38.90}          & \underline{39.51}          & \underline{38.71}          & \underline{39.12}          & \underline{66.2}             \\
\midrule
\rowcolor{ourblue} \multicolumn{8}{c}{\textbf{w/o LLM supervision}} \\
\midrule
CLIP4HOI~\cite{mao2023clip4hoi} & CLIP ViT-L/14 & \checkmark & 35.33 & 33.95 & 35.74 & 34.85 & {66.3} \\
HOIGen~\cite{guo2024unseen} & CLIP ViT-B/16 & \checkmark & 34.84 & 34.52 & 34.94 & 34.74 & - \\
SSGR~\cite{cao2026semantic}                 & CLIP ViT-B/16 & \checkmark                                & 36.86          & 35.74          & 37.20          & 36.47          & 64.2          \\
CMMP~\cite{lei2024exploring} & CLIP ViT-L/14 & \checkmark & {38.14} & {37.75} & {38.25} & {38.00} & 64.0 \\
LAIN~\cite{kim2025LAIN}$^\dagger$                 & CLIP ViT-L/14 & \checkmark                                 & 37.30          & 38.69         & 36.88          & 37.79          & 65.1          \\
LAIN~\cite{kim2025LAIN}$^\dagger$         & DINOtxt ViT-L/16 & \checkmark                            &    36.73 & 35.96 & 36.95 & 36.46          & 63.2          \\
LAIN~\cite{kim2025LAIN}$^\dagger$         & DINOtxt ViT-L/16 &                                & 34.37          & 35.19          & 34.13          & 34.66          & 62.3          \\
\midrule
\textbf{\method} & DINOtxt ViT-L/16 &                                 & \textbf{39.13} & \textbf{39.91} & \textbf{38.89}          & \textbf{39.24} & \textbf{66.5}          \\ 
\bottomrule
\end{tabular}%
}
\vspace{-1em}
\end{table}

\noindent \textbf{Zero-shot results.}
\method achieves state-of-the-art performance across most zero-shot settings on HICO-DET (\cref{tab:main:hicodet_zs}), particularly on UV, UC-RF, UC-NF, and UC, with consistent gains across these splits. The largest improvements are observed in the UV and UC settings, where, for example, \method reaches \textbf{35.58\%} unseen mAP, outperforming the second-best method LAIN by \textbf{+3.53} points in UV and \textbf{+0.33} in UC. Other strong gains are also obtained on UC-RF and UC-NF settings, achieving \textbf{37.79\%} and \textbf{38.11\%}, respectively, demonstrating robust generalization to unseen interaction compositions.
On the UO setting, \method remains competitive but does not surpass methods that leverage additional sources of object-centric features ($\mathbf{\Delta}$): LAIN~\cite{kim2025LAIN} and VDRP~\cite{yang2025visual}. Such features can provide useful cues due to their pretraining on overlapping object classes. We analyze this effect below, showing empirically that incorporating detector-centric features yields noticeable gains for LAIN, while \method already captures object-agnostic interaction cues effectively without relying on such features.
Furthermore, the other best performer on UO, VDRP, benefits from external LLM supervision that enriches object semantics. In contrast, \method does not rely on additional LLM-derived supervision.
To ensure a fair comparison, we additionally evaluate LAIN under the same backbone and feature setting (\ie, without $\mathbf{\Delta}$, see~\supmat\ for the illustration of the usage of $\mathbf{\Delta}$ by the prior methods). Under this setup, \method consistently outperforms all methods across all settings, confirming its effectiveness.

\begin{table}[!t]
\caption{Ablation of \method on HICO-DET dataset under the UV setting. SM stands for soft-masks (\cref{sec:main:method:object_detector_and_soft_mask}), APA for \adaptivepatchmodule (\cref{sec:main:method:adaptive_grounding}), and $\mathcal{I}$ for the \regiontransformermodule (\cref{sec:main:method:interaction_transformer}). Best results in bold.}
\label{tab:main:abl_hicodet_uv}
\resizebox{0.55\columnwidth}{!}{%
\begin{tabular}{@{}ccccccc@{}}
\toprule
\textbf{SM} & \textbf{APA} & \textbf{$\mathcal{I}$} & \textbf{Full} & \textbf{Seen} & \textbf{Unseen} & \textbf{HM} \\ \midrule
\xmark     & \xmark                                                           & \xmark                                                                         & 33.39 & 34.53 & 26.42      & 30.48  \\
\cmark     & \xmark                                                           & \xmark                                                                         & 35.36 & 36.05 &  31.13      & 33.59  \\
\cmark     & \cmark                                                           & \xmark                                                                         & 35.36 & 35.55 & 34.21      & 34.88  \\
\cmark     & \cmark                                                           & \cmark                                                                         & \textbf{38.10}    & \textbf{38.51}    & \textbf{35.58}      & \textbf{37.04}  \\ \bottomrule
\end{tabular}%
}
\end{table}

\begin{table}[!t]
\caption{Effectiveness of soft-mask on the HICO-DET dataset under the UV setting. Best results in bold.}
\label{tab:main:abl_hicodet_masks}
\resizebox{0.55\columnwidth}{!}{%
\begin{tabular}{@{}lcccc@{}}
\toprule
\multicolumn{1}{c}{} & \textbf{Full} & \textbf{Seen} & \textbf{Unseen} & \textbf{HM} \\ \midrule
RoIAlign                 & 35.61 & 36.34 & 31.13            & 33.73 \\
Uniform                 & 35.41 & 35.68 & 33.80             & 34.74 \\
Soft-mask                 & \textbf{38.10}             & \textbf{38.51}             & \textbf{35.58}               & \textbf{37.04}   \\
\bottomrule
\end{tabular}%
}
\vspace{-1em}
\end{table}

\noindent \textbf{Closed-set results.} Although \method is designed for zero-shot HOI detection, we also evaluate it in the closed-set setting. As shown in \cref{tab:main_hicodet_closeset}, \method achieves the best performance among all compared methods, reaching \textbf{39.13} mAP on HICO-DET, including \textbf{39.91} mAP on the Rare split, and \textbf{66.5} $\mathrm{AP}^{\mathrm{S2}}$ on V-COCO.
In particular, \method outperforms the strongest competing methods that rely on additional supervision. Compared to VDRP~\cite{yang2025visual}, which leverages LLM guidance and detector-centric features, \method improves performance by \textbf{+0.40} mAP on the Rare split. Furthermore, compared to LAIN~\cite{kim2025LAIN} under the same DINOtxt-based setting (without detector-centric features $\Delta$), \method achieves gains of \textbf{+4.76} mAP on Full, \textbf{+4.72} mAP on Rare, and \textbf{+4.58} HM.
This shows that improved visual grounding remains beneficial even when all interaction classes are observed during training. For completeness, we also report an adapted LAIN baseline under the same DINOtxt-based setting. The performance gap remains consistent, with \method outperforming LAIN across all metrics, confirming that our part-aware interaction representation provides advantages beyond the zero-shot regime.

\subsection{Ablation Studies}

\noindent \textbf{Component analysis.} We progressively enable each component of \method to analyze its individual contributions (\cref{tab:main:abl_hicodet_uv}). The baseline (first row) corresponds to a setting where the proposed soft masks, \adaptivepatchmodule, and \regiontransformermodule are replaced with standard RoIAlign features and a vanilla transformer.
As components are added, both Unseen mAP and HM improve consistently, indicating improved zero-shot generalization.
Notably, incorporating the soft-mask module (SM) yields a noticeable improvement of \textbf{+4.71} Unseen mAP and \textbf{+3.11} HM, while also improving Seen performance, indicating that patch-level grounding provides stronger spatial alignment than standard RoI features.
Introducing \adaptivepatchmodule (APA) further boosts Unseen mAP by \textbf{+3.08} and HM by \textbf{+1.29}, with negligible impact on Seen performance. This suggests that part decomposition primarily enhances compositional generalization rather than fitting seen interactions.
Finally, incorporating the \regiontransformermodule ($\mathcal{I}$) leads to the strongest overall performance, improving Full mAP by \textbf{+2.74}, Seen mAP by \textbf{+2.96}, and Unseen mAP by \textbf{+1.37}, highlighting the importance of integrating localized interaction cues with global context.
Overall, the three components are complementary: soft masks improve spatial grounding, \adaptivepatchmodule strengthens unseen generalization, and the \regiontransformermodule provides contextual refinement. 
We conduct these ablations on the UV setting, which directly evaluates generalization to novel verbs, following prior work~\cite{kim2025LAIN}. A similar trend is observed across zero-shot settings: UO, UC-RF, UC-NF, and UC. Corresponding results are provided in \supmat.

\noindent \textbf{Soft-mask feature extraction formulation.} \cref{tab:main:abl_hicodet_masks} evaluates the impact of the proposed soft-mask by comparing it with alternative feature pooling strategies. We consider three variants: (1) RoIAlign~\cite{he2017mask}, (2) uniform average pooling over patch features, and (3) the proposed semantically weighted soft-mask (\cref{sec:main:method:object_detector_and_soft_mask}).
Replacing RoIAlign with a uniform mask already improves zero-shot generalization, yielding a gain of \textbf{+2.67\%} Unseen mAP and \textbf{+1.01\%} HM. However, this comes at the cost of a \textbf{-0.66\%} drop in Seen performance, indicating that uniform aggregation alone is insufficient to preserve the most informative features.
In contrast, the proposed soft-mask achieves the best performance across all metrics, improving over RoIAlign by \textbf{+2.49\%} Full, \textbf{+2.17\%} Seen, \textbf{+4.45\%} Unseen, and \textbf{+3.31\%} HM. This demonstrates that semantically weighting patch features is crucial for interaction modeling, as it emphasizes interaction-relevant regions while suppressing less informative areas.
Overall, these results highlight a clear progression: while coarse spatial selection (uniform) already benefits generalization, incorporating semantic weighting is essential to fully exploit fine-grained interaction cues.

\begin{table}[!t]
\caption{Effectiveness of multimodal embeddings on HICO-DET under the UV setting. Best results in bold.}
\label{tab:main:abl_hicodet_prompts}
\resizebox{0.8\columnwidth}{!}{%
\begin{tabular}{@{}lcccc@{}}
\toprule
\textbf{} & \textbf{Full} & \textbf{Seen} & \textbf{Unseen} & \textbf{HM} \\ \midrule
HO token ($\tilde{\mathbf{h}}_{ho}$) & 35.49 & 36.21 & 31.07 & 33.64 \\
~+ global ($\bar{\mathbf{e}}_{g}$) & 36.18 & 37.05 & 30.79 & 33.92 \\
~+ global ($\bar{\mathbf{e}}_{g}$) + mask-guided ($\bar{\mathbf{e}}_{m}$) & \textbf{38.10} & \textbf{38.51} & \textbf{35.58} & \textbf{37.04} \\
\bottomrule
\end{tabular}%
}
\end{table}

\noindent \textbf{On the effectiveness of masked embeddings.} We analyze the contribution of the visual components used to construct the final multimodal embedding within the \regiontransformermodule (\cref{sec:main:method:interaction_transformer}). As shown in \cref{tab:main:abl_hicodet_prompts}, we consider three configurations: (a) using only the human-object token ($\tilde{\mathbf{h}}_{ho}$), (b) augmenting it with the global branch ($\bar{\mathbf{e}}_{g}$), and (c) the full model, with the global ($\bar{\mathbf{e}}_{g}$) and mask-guided ($\bar{\mathbf{e}}_{m}$) branches. 
Adding the global branch improves Full and Seen performance by \textbf{+0.69\%} and \textbf{+0.84\%}, respectively, but slightly reduces Unseen mAP by \textbf{-0.28\%}. This suggests that global context provides useful scene-level cues, but may also introduce irrelevant information that hinders generalization to unseen interactions.
In contrast, incorporating the mask-guided branch (on top of the global branch) yields substantial improvements across all metrics, boosting Full by \textbf{+1.92\%}, Seen by \textbf{+1.46\%}, Unseen by \textbf{+4.79\%}, and HM by \textbf{+3.12\%}. This highlights the importance of localized, interaction-aware features for zero-shot HOI recognition.
Overall, the two branches are complementary: the global branch captures coarse contextual information, while the mask-guided branch focuses on interaction-relevant regions, and their combination enables robust and fine-grained interaction modeling.

\begin{figure}[!t]
    \centering
    \includegraphics[width=0.9\columnwidth]{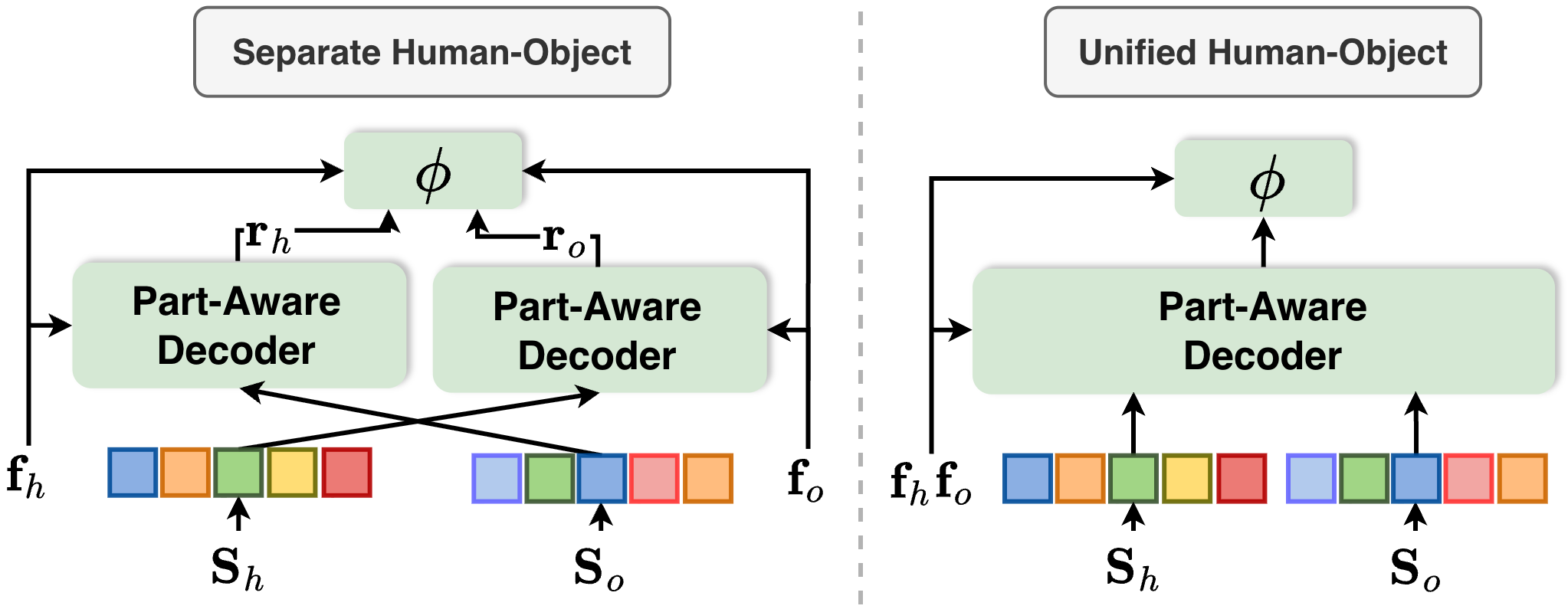}
    \caption{Visualization of the part-level attention module with separate or unified human and object parts decoder.}
    \vspace{-1em}
    \label{fig:main:part_level_attention}
\end{figure}

\begin{table}[!t]
\caption{Comparison between unified and separate part-level attention on the HICO-DET dataset under the UV setting.}
\label{tab:main:abl_hicodet_partlevel}
\resizebox{0.8\columnwidth}{!}{%
\begin{tabular}{@{}lcccc@{}}
\toprule
\multicolumn{1}{c}{}                              & \textbf{Full}  & \textbf{Seen}  & \textbf{Unseen} & \textbf{HM}    \\ \midrule
Unified human/object          &  36.36 & 37.01 & 32.34   & 34.66          \\
Separate human/object    & \textbf{38.10}          & \textbf{38.51}          & \textbf{35.58}           & \textbf{37.04}          \\
\bottomrule
\end{tabular}%
}
\vspace{-1em}
\end{table}

\begin{figure*}[!t]
    \centering
    \includegraphics[width=0.9\textwidth]{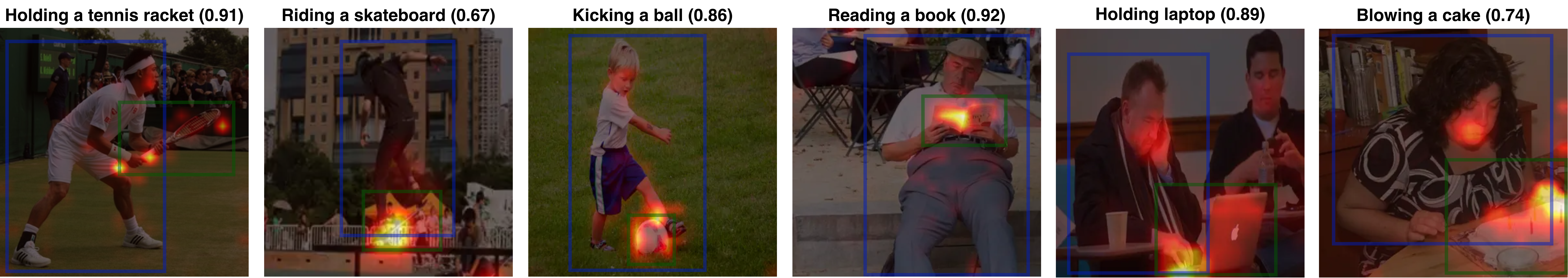}
    \caption{Attention maps of the interaction transformer $\mathcal{I}$ in \method. Warmer colors indicate higher attention to visual patches. Blue and green boxes denote the human and object detections, respectively.}
    \label{fig:main:qualitatives}
\end{figure*}

\noindent \textbf{Separate versus unified part-level attention.}
We compare our proposed part-level attention (\cref{sec:main:method:adaptive_grounding}) with an alternative ``unified'' approach where the human $\mathbf{f}_h$ and object $\mathbf{f}_o$ features are concatenated into a single query sequence (\cref{fig:main:part_level_attention}, right), and the human $\mathbf{S}_h$ and object $\mathbf{S}_o$ sub-parts are concatenated into a single key/value sequence. As shown in \cref{tab:main:abl_hicodet_partlevel}, modeling the human and object separately yields consistently better results, improving Full mAP by \textbf{+1.74\%}, Seen mAP by \textbf{+1.50\%}, Unseen mAP by \textbf{+3.24\%}, and HM by \textbf{+2.38\%}. The gains are especially pronounced on the Unseen split, with an improvement of \textbf{+3.24\%}, which is more than twice the gain observed on Seen (\textbf{+1.50\%}). This suggests that separate attention better supports generalization to novel interactions. 
We attribute this behavior to the asymmetric nature of HOIs: the human should attend to object parts, while the object should attend to human parts. In contrast, the unified formulation mixes the two roles into a single attention space, which weakens entity-specific conditioning and makes it harder to localize the interaction-relevant regions.

\begin{table}[!t]
\caption{Performance of \method and LAIN with DINOtxt and detector features ($\Delta$) on HICO-DET under the UO setting.}
\label{tab:main:abl_hicodet_detr}
\resizebox{0.6\columnwidth}{!}{%
\begin{tabular}{@{}llcccc@{}}
\toprule
\multicolumn{1}{c}{}                             & $\Delta$ & \textbf{Full}  & \textbf{Seen}  & \textbf{Unseen} & \textbf{HM}    \\ \midrule
\multicolumn{1}{l}{\multirow{2}{*}{LAIN~\cite{kim2025LAIN}}} & \checkmark      &          {36.01}                             & {34.72}                             & {38.11}                               & {36.41}          \\
\multicolumn{1}{l}{}                      &                 & 34.85          & 34.38          & 37.18           & 35.78          \\
\midrule
\multicolumn{1}{l}{\multirow{2}{*}{\textbf{\method (Ours)}}}           & \checkmark                & \textbf{36.84}          & \textbf{35.93}          & \textbf{41.42}           & \textbf{38.67}          \\
\multicolumn{1}{l}{}                                          &                & 36.62 & 35.92 & 40.13  & 38.03 \\
\bottomrule
\end{tabular}%
}
\vspace{-1em}
\end{table}

\noindent \textbf{Effect of detector priors in the UO setting.}
We analyze the role of object detector features ($\mathbf{\Delta}$) in the UO setting, where generalization depends on recognizing object categories not observed during training. In this regime, detector features provide auxiliary cues due to pretraining on overlapping object classes.
As shown in \cref{tab:main:abl_hicodet_detr}, incorporating $\mathbf{\Delta}$ leads to moderate improvements in Unseen performance for both methods. However, \method consistently outperforms LAIN both with and without $\mathbf{\Delta}$. Notably, even without $\mathbf{\Delta}$, \method surpasses LAIN equipped with $\mathbf{\Delta}$ by \textbf{+2.02\%} Unseen mAP and \textbf{+1.62\%} HM.
These results indicate that, while $\mathbf{\Delta}$ can provide complementary information, the proposed part-aware interaction representation already captures transferable interaction cues effectively, enabling stronger generalization to unseen object categories without relying on detector-centric features.

\begin{table}[!t]
\caption{Comparison between our proposed soft-mask and SAM2 on the HICO-DET dataset under all zero-shot settings.}
\label{tab:main:abl_hicodet_sam2}
\resizebox{0.6\columnwidth}{!}{%
\begin{tabular}{@{}ll|cccc@{}}
\toprule
\multicolumn{1}{c}{}    & \multicolumn{1}{c}{} & \multicolumn{1}{c}{\textbf{Full}} & \multicolumn{1}{c}{\textbf{Seen}} & \multicolumn{1}{c}{\textbf{Unseen}} & \multicolumn{1}{c}{\textbf{HM}} \\ \midrule
\multirow{2}{*}{UV} & SAM2                   & 37.57 & 38.20 & 33.68                               & 35.94                           \\
                    & Soft-mask                       & \textbf{38.10}                             & \textbf{38.51}                             & \textbf{35.58}                               & \textbf{37.04}                           \\
\midrule
\multirow{2}{*}{UO} & SAM2                   & 36.61                             & \textbf{36.38}                             & 37.70                               & 37.04                           \\
                    & Soft-mask                       & \textbf{36.62}                             & 35.92                             & \textbf{40.13}                               & \textbf{38.03}                           \\
\midrule
\multirow{2}{*}{UC-RF} & SAM2                   & 37.90                             & 37.95                             & 37.69                               & 37.82                           \\
                    & Soft-mask                       & \textbf{38.39}                             & \textbf{38.54}                             & \textbf{37.79}                               & \textbf{38.17}                           \\
\midrule
\multirow{2}{*}{UC-NF} & SAM2                   & 36.12                             & 35.93                             & 36.84                               & 36.39                           \\
                    & Soft-mask                       & \textbf{36.58}                             & \textbf{36.20}                             & \textbf{38.11}                               & \textbf{37.15}                           \\
\midrule
\multirow{2}{*}{UC} & SAM2                   & 35.13 & 	35.25 &	34.25 &	34.54                           \\
                    & Soft-mask                       & \textbf{36.88}	& \textbf{37.41} &	\textbf{34.79}	& \textbf{36.10}                           \\
\bottomrule
\end{tabular}%
}
\vspace{-1.5em}
\end{table}

\noindent \textbf{Comparison between soft masks and SAM2.}
We further compare our proposed soft-mask formulation (\cref{sec:main:method:object_detector_and_soft_mask}) against binary masks generated by SAM2~\cite{ravi2024sam}, a strong segmentation model, using DETR bounding boxes as prompts (see \supmat~for additional details on the implementation of SAM2). Despite SAM2 providing high-quality binary masks,
our soft-mask consistently performs better across all zero-shot settings. In particular, it improves Full mAP in UV, UC-RF, UC-NF, UC, and yields especially large gains on the Unseen split (UO \textbf{+2.43\%}, UV \textbf{+1.90\%}, UC-NF \textbf{+1.27\%}). Moreover, soft-mask achieves a higher harmonic mean (HM) in all settings (UO \textbf{+0.99\%}, UV \textbf{+1.10\%}, and UC \textbf{+1.56\%}), indicating a better balance between Seen and Unseen performance. We argue that, in our setting, binary masks (like the ones generated by SAM2) impose an overly strict separation between foreground and background. In contrast, soft masks preserve evidence around boundaries and contact regions, which proves beneficial for modeling fine-grained interactions.

\subsection{Qualitative Results}
\label{sec:main:qualitatives}
We present the attention maps of the interaction transformer $\mathcal{I}$ in~\cref{fig:main:qualitatives} to analyze how \method localizes interaction cues. Across diverse HOI categories, the attention consistently focuses on localized regions that are directly relevant to the predicted interaction, such as hand-racket contact area in $\langle$\textit{human}, \textit{holding}, \textit{tennis racket}$\rangle$, ball-foot in $\langle$\textit{human}, \textit{kicking}, \textit{ball}$\rangle$, hands in $\langle$\textit{human}, \textit{holding}, \textit{laptop}$\rangle$, or mouth in $\langle$\textit{human}, \textit{blowing}, \textit{cake}$\rangle$.
These visualizations indicate that \method effectively identifies interaction-critical regions, rather than relying on coarse object-level features. This behavior supports our design choice of combining soft-mask grounding and part-level attention, which enables fine-grained reasoning over HOIs.

\section{Conclusion}
\label{sec:main:conclusion}
We present \method, a zero-shot HOI detection framework that models interactions from dense patch-level representations. Moving beyond detector-centric features, our approach captures fine-grained spatial relationships through soft-mask grounding, adaptive part-level attention, and a region-aware interaction transformer, enabling the model to focus on interaction-critical regions while preserving contextual information.
Extensive experiments show that \method achieves state-of-the-art performance across standard zero-shot settings, with particularly strong results on unseen interactions. Ablation studies further demonstrate that each component contributes to improved generalization, with the largest gains stemming from semantically grounded patch representations and part-aware reasoning.
Future work will explore richer interaction reasoning in the zero-shot setting, including training-free approaches that leverage broader semantic context beyond pairwise human-object regions. We also plan to extend the framework to video-based HOI understanding.

\clearpage
\section*{Acknowledgements}
We thank the EuroHPC initiative for the availability of high performance computing resources. 
This work was supported by the EU Horizon projects ELIAS (No. 101120237), ELLIOT (No. 101214398), TURING (No. 101215032), IAMI (No. 101168272), and by the Autonomous Province of Trento under Provincial Law No. 6/2023, as part of the VisRec project (CUP C69J25000880001).
FT acknowledges travel support from ELIAS (GA No. 101120237).

\bibliographystyle{ACM-Reference-Format}
\balance
\bibliography{main}

@String{Computer = "{IEEE} Computer" }

@String{Springer = "Springer-Verlag" }

@article{simeoni2025dinov3,
  title={Dinov3},
  author={Sim{\'e}oni, Oriane and Vo, Huy V and Seitzer, Maximilian and Baldassarre, Federico and Oquab, Maxime and Jose, Cijo and Khalidov, Vasil and Szafraniec, Marc and Yi, Seungeun and Ramamonjisoa, Micha{\"e}l and others},
  journal={arXiv preprint arXiv:2508.10104},
  year={2025}
}

@inproceedings{radford2021learning,
  title={Learning transferable visual models from natural language supervision},
  author={Radford, Alec and Kim, Jong Wook and Hallacy, Chris and Ramesh, Aditya and Goh, Gabriel and Agarwal, Sandhini and Sastry, Girish and Askell, Amanda and Mishkin, Pamela and Clark, Jack and others},
  booktitle={International conference on machine learning},
  pages={8748--8763},
  year={2021},
  organization={PmLR}
}

@inproceedings{geng2025horp,
  title={Horp: Human-object relation priors guided hoi detection},
  author={Geng, Pei and Yang, Jian and Zhang, Shanshan},
  booktitle={Proceedings of the Computer Vision and Pattern Recognition Conference},
  pages={25325--25335},
  year={2025}
}

@article{lei2024ez,
  title={Ez-hoi: Vlm adaptation via guided prompt learning for zero-shot hoi detection},
  author={Lei, Qinqian and Wang, Bo and Tan, Robby},
  journal={Advances in Neural Information Processing Systems},
  volume={37},
  pages={55831--55857},
  year={2024}
}

@article{yang2025visual,
  title={Visual Diversity and Region-aware Prompt Learning for Zero-shot HOI Detection},
  author={Yang, Chanhyeong and Song, Taehoon and Park, Jihwan and Kim, Hyunwoo J},
  journal={arXiv preprint arXiv:2510.25094},
  year={2025}
}

@article{cao2026semantic,
  title={Semantic-Spatial Guided Reasoning for Human-Object Interaction Detection},
  author={Cao, Ping and Zhang, Chunjie and Zheng, Xiaolong and Zhao, Yao},
  journal={IEEE Signal Processing Letters},
  volume={33},
  pages={551--555},
  year={2026},
  publisher={IEEE}
}

@article{shi2000normalized,
  title={Normalized cuts and image segmentation},
  author={Shi, Jianbo and Malik, Jitendra},
  journal={IEEE Transactions on pattern analysis and machine intelligence},
  volume={22},
  number={8},
  pages={888--905},
  year={2000},
  publisher={Ieee}
}

@inproceedings{Chao2018,
  author       = {Yu{-}Wei Chao and
                  Yunfan Liu and
                  Xieyang Liu and
                  Huayi Zeng and
                  Jia Deng},
  title        = {Learning to Detect Human-Object Interactions},
  booktitle    = {{IEEE} Winter Conf. Appl. Comp. Vis. ({WACV})},
  pages        = {381--389},
  year         = {2018},
  address = {Lake Tahoe, NV, USA},
  publisher    = {{IEEE} Computer Society},
}

@article{Gupta2015,
  author       = {Saurabh Gupta and
                  Jitendra Malik},
  title        = {Visual Semantic Role Labeling},
  journal      = {CoRR},
  volume       = {abs/1505.04474},
  year         = {2015},
  archivePrefix = {arXiv},
  pages = {1--11}
}

@inproceedings{hou2021detecting,
  title={Detecting human-object interaction via fabricated compositional learning},
  author={Hou, Zhi and Yu, Baosheng and Qiao, Yu and Peng, Xiaojiang and Tao, Dacheng},
  booktitle={{IEEE} Conf. Comput. Vis. Pattern Recog. ({CVPR})},
  pages={14646--14655},
  year={2021},
  address = {Virtual},
  publisher    = {Computer Vision Foundation / {IEEE}}
}

@misc{yang2024alignedcutvisualconceptsdiscovery,
      title={AlignedCut: Visual Concepts Discovery on Brain-Guided Universal Feature Space}, 
      author={Huzheng Yang and James Gee and Jianbo Shi},
      year={2024},
      eprint={2406.18344},
      archivePrefix={arXiv},
      primaryClass={cs.CV}, 
}

@inproceedings{lei2024exploring,
  title={Exploring conditional multi-modal prompts for zero-shot hoi detection},
  author={Lei, Ting and Yin, Shaofeng and Peng, Yuxin and Liu, Yang},
  booktitle={European Conference on Computer Vision},
  pages={1--19},
  year={2024},
  organization={Springer}
}

@article{mao2023clip4hoi,
  title={Clip4hoi: towards adapting clip for practical zero-shot hoi detection},
  author={Mao, Yunyao and Deng, Jiajun and Zhou, Wengang and Li, Li and Fang, Yao and Li, Houqiang},
  journal={Advances in Neural Information Processing Systems},
  volume={36},
  pages={45895--45906},
  year={2023}
}

@inproceedings{tonini2025dynamic,
  title={Dynamic Scoring with Enhanced Semantics for Training-Free Human-Object Interaction Detection},
  author={Tonini, Francesco and Vaquero, Lorenzo and Conti, Alessandro and Beyan, Cigdem and Ricci, Elisa},
  booktitle={Proceedings of the 33rd ACM International Conference on Multimedia},
  pages={2801--2810},
  year={2025}
}

@inproceedings{kim2025LAIN,
  title={Locality-aware zero-shot human-object interaction detection},
  author={Kim, Sanghyun and Jung, Deunsol and Cho, Minsu},
  booktitle={Proceedings of the Computer Vision and Pattern Recognition Conference},
  pages={20190--20200},
  year={2025}
}

@inproceedings{Carion2020,
  author       = {Nicolas Carion and
                  Francisco Massa and
                  Gabriel Synnaeve and
                  Nicolas Usunier and
                  Alexander Kirillov and
                  Sergey Zagoruyko},
  title        = {End-to-End Object Detection with Transformers},
  booktitle    = {European Conf. Comput. Vis. ({ECCV})},
  pages        = {213--229},
  year         = {2020},
  address = {Glasgow, UK},
  volume       = {12346},
  publisher    = {Springer}
}

@inproceedings{Kim2021,
  author       = {Bumsoo Kim and
                  Junhyun Lee and
                  Jaewoo Kang and
                  Eun{-}Sol Kim and
                  Hyunwoo J. Kim},
  title        = {{HOTR:} End-to-End Human-Object Interaction Detection With Transformers},
  booktitle    = {{IEEE} Conf. Comput. Vis. Pattern Recog. ({CVPR})},
  pages        = {74--83},
  year         = {2021},
  address = {Virtual},
  publisher    = {Computer Vision Foundation / {IEEE}}
}

@inproceedings{Bansal2020,
  author    = {Ankan Bansal and Sai Saketh Rambhatla and Abhinav Shrivastava and Rama Chellappa},
  title     = {Detecting Human-Object Interactions via Functional Generalization},
  booktitle = {Proceedings of the AAAI Conference on Artificial Intelligence (AAAI)},
  volume    = {34},
  pages     = {10460--10469},
  year      = {2020}
}

@inproceedings{VCL,
  author    = {Zhi Hou and Xiaojiang Peng and Yu Qiao and Dacheng Tao},
  title     = {Visual Compositional Learning for Human-Object Interaction Detection},
  booktitle = {Proceedings of the European Conference on Computer Vision (ECCV)},
  pages     = {584--600},
  year      = {2020}
}

@inproceedings{zhang2022exploring,
  title={Exploring structure-aware transformer over interaction proposals for human-object interaction detection},
  author={Zhang, Yong and Pan, Yingwei and Yao, Ting and Huang, Rui and Mei, Tao and Chen, Chang-Wen},
  booktitle={Proceedings of the IEEE/CVF Conference on Computer Vision and Pattern Recognition},
  pages={19548--19557},
  year={2022}
}

@inproceedings{Kim2022,
  author       = {Bumsoo Kim and
                  Jonghwan Mun and
                  Kyoung{-}Woon On and
                  Minchul Shin and
                  Junhyun Lee and
                  Eun{-}Sol Kim},
  title        = {{MSTR:} Multi-Scale Transformer for End-to-End Human-Object Interaction
                  Detection},
  booktitle    = {{IEEE} Conf. Comput. Vis. Pattern Recog. ({CVPR})},
  pages        = {19556--19565},
  year         = {2022},
  address = {New Orleans, LA, USA},
  publisher    = {{IEEE}}
}

@inproceedings{Tamura2021,
  author       = {Masato Tamura and
                  Hiroki Ohashi and
                  Tomoaki Yoshinaga},
  title        = {{QPIC:} Query-Based Pairwise Human-Object Interaction Detection With
                  Image-Wide Contextual Information},
  booktitle    = {{IEEE} Conf. Comput. Vis. Pattern Recog. ({CVPR})},
  pages        = {10410--10419},
  year         = {2021},
  address = {Virtual},
  publisher    = {Computer Vision Foundation / {IEEE}}
}

@inproceedings{Tu2022,
  author       = {Danyang Tu and
                  Xiongkuo Min and
                  Huiyu Duan and
                  Guodong Guo and
                  Guangtao Zhai and
                  Wei Shen},
  title        = {Iwin: Human-Object Interaction Detection via Transformer with Irregular
                  Windows},
  booktitle    = {European Conf. Comput. Vis. ({ECCV})},
  pages        = {87--103},
  year         = {2022},
  address    = {Tel Aviv, Israel},
  publisher    = {Springer}
}

@inproceedings{Zhong2022,
  author       = {Xubin Zhong and
                  Changxing Ding and
                  Zijian Li and
                  Shaoli Huang},
  title        = {Towards Hard-Positive Query Mining for DETR-Based Human-Object Interaction
                  Detection},
  booktitle    = {European Conf. Comput. Vis. ({ECCV})},
  pages        = {444--460},
  year         = {2022},
  address    = {Tel Aviv, Israel},
  publisher    = {Springer}
}

@inproceedings{Zhang2022,
  author       = {Frederic Z. Zhang and
                  Dylan Campbell and
                  Stephen Gould},
  title        = {Efficient Two-Stage Detection of Human-Object Interactions with a
                  Novel Unary-Pairwise Transformer},
  booktitle    = {{IEEE} Conf. Comput. Vis. Pattern Recog. ({CVPR})},
  pages        = {20072--20080},
  year         = {2022},
  address = {New Orleans, LA, USA},
  publisher    = {{IEEE}},
}

@inproceedings{Gao2020,
  author       = {Chen Gao and
                  Jiarui Xu and
                  Yuliang Zou and
                  Jia{-}Bin Huang},
  title        = {{DRG:} Dual Relation Graph for Human-Object Interaction Detection},
  booktitle    = {European Conf. Comput. Vis. ({ECCV})},
  pages        = {696--712},
  year         = {2020},
  address = {Glasgow, UK},
  volume       = {12357},
  publisher    = {Springer},
}

@inproceedings{Zhang2021,
  author       = {Frederic Z. Zhang and
                  Dylan Campbell and
                  Stephen Gould},
  title        = {Spatially Conditioned Graphs for Detecting Human-Object Interactions},
  booktitle    = {{IEEE} Int. Conf. Comput. Vis. ({ICCV})},
  pages        = {13299--13307},
  year         = {2021},
  address    = {Montreal, QC, Canada},
  publisher    = {{IEEE}}
}

@inproceedings{Lei2023,
  author       = {Ting Lei and
                  Fabian Caba and
                  Qingchao Chen and
                  Hailin Jin and
                  Yuxin Peng and
                  Yang Liu},
  title        = {Efficient Adaptive Human-Object Interaction Detection with Concept-guided
                  Memory},
  booktitle    = {{IEEE} Int. Conf. Comput. Vis. ({ICCV})},
  pages        = {6457--6467},
  year         = {2023},
  address    = {Paris, France},
  publisher    = {{IEEE}}
}

@ARTICLE{9583890,
  author={Yang, Xu and Zhang, Hanwang and Cai, Jianfei},
  journal={{IEEE} Trans. Pattern Anal. Mach. Intell.}, 
  title={Deconfounded Image Captioning: A Causal Retrospect}, 
  year={2023},
  volume={45},
  number={11},
  pages={12996-13010}
}

@InProceedings{Seo_2022_CVPR,
    author    = {Seo, Paul Hongsuck and Nagrani, Arsha and Arnab, Anurag and Schmid, Cordelia},
    title     = {End-to-End Generative Pretraining for Multimodal Video Captioning},
    booktitle = {{IEEE} Conf. Comput. Vis. Pattern Recog. ({CVPR})},
    year      = {2022},
    pages     = {17959-17968},
  address = {New Orleans, LA, USA},
  publisher    = {{IEEE}},
}

@inproceedings{chen2024asynchronous,
  title={Asynchronous large language model enhanced planner for autonomous driving},
  author={Chen, Yuan and Ding, Zi-han and Wang, Ziqin and Wang, Yan and Zhang, Lijun and Liu, Si},
  booktitle={European Conf. Comput. Vis. ({ECCV})},
  pages={22--38},
  year={2024},
  address = {Milan, Italy},
  volume       = {15094},
  publisher    = {Springer},
}

@article{li2024ripple,
  author       = {Huadong Li and
                  Ying Wei and
                  Shuailei Ma and
                  Mingyu Chen and
                  Ge Li},
  title        = {Ripple Transformer: {A} Human-Object Interaction Backbone and a New
                  Prediction Strategy for Smart Surveillance Devices},
  journal      = {{IEEE} Trans. Consumer Electron.},
  volume       = {70},
  number       = {1},
  pages        = {2257--2268},
  year         = {2024}
}

@inproceedings{Gupta2019,
  author       = {Tanmay Gupta and
                  Alexander G. Schwing and
                  Derek Hoiem},
  title        = {No-Frills Human-Object Interaction Detection: Factorization, Layout
                  Encodings, and Training Techniques},
  booktitle    = {{IEEE} Int. Conf. Comput. Vis. ({ICCV})},
  pages        = {9676--9684},
  year         = {2019}
}

@inproceedings{Liu2020b,
  author       = {Ye Liu and
                  Junsong Yuan and
                  Chang Wen Chen},
  title        = {ConsNet: Learning Consistency Graph for Zero-Shot Human-Object Interaction
                  Detection},
  booktitle    = {{ACM} Multimedia ({ACMMM})},
  pages        = {4235--4243},
  year         = {2020},
  address    = {Seattle, WA, USA},
  publisher    = {{ACM}}
}

@inproceedings{Ulutan2020,
  author       = {Oytun Ulutan and
                  A. S. M. Iftekhar and
                  B. S. Manjunath},
  title        = {VSGNet: Spatial Attention Network for Detecting Human Object Interactions
                  Using Graph Convolutions},
  booktitle    = {{IEEE} Conf. Comput. Vis. Pattern Recog. ({CVPR})},
  pages        = {13614--13623},
  year         = {2020},
  address    = {Seattle, WA, USA},
  publisher    = {Computer Vision Foundation / {IEEE}}
}

@inproceedings{zhang2023exploring,
  title={Exploring predicate visual context in detecting of human-object interactions},
  author={Zhang, Frederic Z and Yuan, Yuhui and Campbell, Dylan and Zhong, Zhuoyao and Gould, Stephen},
  booktitle={{IEEE} Int. Conf. Comput. Vis. ({ICCV})},
  pages={10411--10421},
  year={2023},
  address    = {Paris, France},
  publisher    = {{IEEE}}
}

@inproceedings{zhang2021mining,
  author       = {Aixi Zhang and
                  Yue Liao and
                  Si Liu and
                  Miao Lu and
                  Yongliang Wang and
                  Chen Gao and
                  Xiaobo Li},
  title        = {Mining the Benefits of Two-stage and One-stage {HOI} Detection},
  booktitle    = {Adv. Neural Inf. Process. Syst. ({NeurIPS})},
  pages        = {17209--17220},
  year         = {2021},
  volume = {34},
  address = {Virtual},
  publisher = {Curran Associates, Inc.}
}

@article{ravi2024sam,
  title={Sam 2: Segment anything in images and videos},
  author={Ravi, Nikhila and Gabeur, Valentin and Hu, Yuan-Ting and Hu, Ronghang and Ryali, Chaitanya and Ma, Tengyu and Khedr, Haitham and R{\"a}dle, Roman and Rolland, Chloe and Gustafson, Laura and others},
  journal={arXiv preprint arXiv:2408.00714},
  year={2024}
}

@article{herdade2019image,
  title={Image captioning: Transforming objects into words},
  author={Herdade, Simao and Kappeler, Armin and Boakye, Kofi and Soares, Joao},
  journal={Advances in neural information processing systems},
  volume={32},
  year={2019}
}

@inproceedings{lin2014microsoft,
  title={Microsoft coco: Common objects in context},
  author={Lin, Tsung-Yi and Maire, Michael and Belongie, Serge and Hays, James and Perona, Pietro and Ramanan, Deva and Doll{\'a}r, Piotr and Zitnick, C Lawrence},
  booktitle={European conference on computer vision},
  pages={740--755},
  year={2014},
  organization={Springer}
}

@article{loshchilov2017decoupled,
  title={Decoupled weight decay regularization},
  author={Loshchilov, Ilya and Hutter, Frank},
  journal={arXiv preprint arXiv:1711.05101},
  year={2017}
}

@article{dosovitskiy2020image,
  title={An image is worth 16x16 words: Transformers for image recognition at scale},
  author={Dosovitskiy, Alexey and Beyer, Lucas and Kolesnikov, Alexander and Weissenborn, Dirk and Zhai, Xiaohua and Unterthiner, Thomas and Dehghani, Mostafa and Minderer, Matthias and Heigold, Georg and Gelly, Sylvain and others},
  journal={arXiv preprint arXiv:2010.11929},
  year={2020}
}

@inproceedings{lin2017focal,
  title={Focal loss for dense object detection},
  author={Lin, Tsung-Yi and Goyal, Priya and Girshick, Ross and He, Kaiming and Doll{\'a}r, Piotr},
  booktitle={Proceedings of the IEEE international conference on computer vision},
  pages={2980--2988},
  year={2017}
}

@inproceedings{guo2024unseen,
  title={Unseen no more: Unlocking the potential of clip for generative zero-shot hoi detection},
  author={Guo, Yixin and Liu, Yu and Li, Jianghao and Wang, Weimin and Jia, Qi},
  booktitle={Proceedings of the 32nd ACM International Conference on Multimedia},
  pages={1711--1720},
  year={2024}
}

@inproceedings{he2017mask,
  title={Mask r-cnn},
  author={He, Kaiming and Gkioxari, Georgia and Doll{\'a}r, Piotr and Girshick, Ross},
  booktitle={Proceedings of the IEEE international conference on computer vision},
  pages={2961--2969},
  year={2017}
}

\newpage
\setcounter{page}{1}
\appendix
\maketitlesupplementary

In this supplementary material, we provide additional details and analyses that complement the main paper. We first describe additional implementation choices underlying \method, including backbone architectures, training setup, and the adaptation of LAIN for a fair comparison under our unified framework (\cref{sec:supp:implementation}). We then report further experimental results across the remaining zero-shot settings, together with an additional study on the effect of the number of decomposed parts (\cref{sec:supp_experiments}). Finally, we present additional qualitative visualizations to illustrate how the proposed interaction module localizes the visual evidence relevant to human-object interactions (\cref{sec:supp:qualitative}).

\section{Additional implementation details}
\label{sec:supp:implementation}
We use DETR~\cite{Carion2020} with a ResNet-50 backbone as the frozen object detector $\mathcal{OD}$, and DINOtxt~\cite{simeoni2025dinov3} with a ViT-L/16 backbone for the vision $\mathcal{V}$ and text $\mathcal{T}$ encoders. The \regiontransformermodule is implemented as a Transformer with 2 layers, 16 attention heads, and hidden dimensionality 1024. We use the default preprocessing pipelines of the pretrained DETR and DINOtxt models, together with standard data augmentations during training.
For the human branch, we use the text embedding of the class ``\texttt{person}'', while for the object branch, we use the text embedding of the detected object category $\mathbf{c}_o$. \\

\begin{figure}[!t]
    \centering
    \includegraphics[width=\columnwidth]{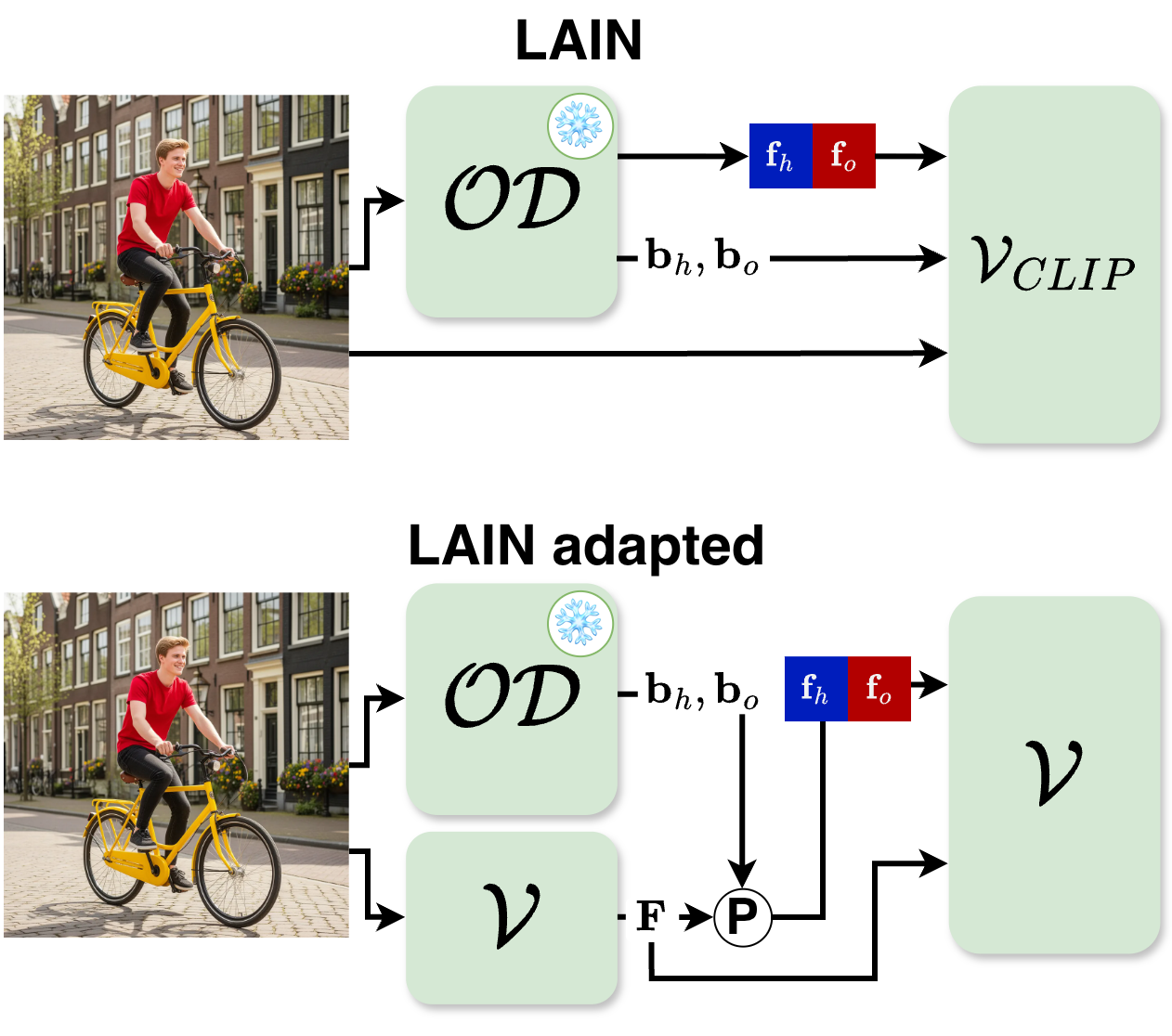}
    \caption{Comparison between the original LAIN~\cite{kim2025LAIN} and our adapted version. In LAIN, DETR ($\mathcal{OD}$) provides both the detections ($\mathbf{b}_h$ and $\mathbf{b}_o$) and the human/object features ($\mathbf{f}_h$ and $\mathbf{f}_o$) used to construct the interaction representation, which is then processed jointly with the image by a CLIP-based visual encoder ($\mathcal{V}_{CLIP}$). In our adaptation, we replace CLIP with DINOtxt ($\mathcal{V}$) and use DETR only for bounding box prediction. Human and object features are obtained by average pooling (\protect\textcircled{P}) the DINOtxt patch features within the corresponding detected boxes.}
    \label{fig:supp:lain_adaptation}
\end{figure}

\noindent \textbf{Adaptation of LAIN~\cite{kim2025LAIN}}.
To ensure a fair comparison between \method and prior state-of-the-art approaches, we modify LAIN~\cite{kim2025LAIN} in two ways, as shown in \cref{fig:supp:lain_adaptation}. First, we replace LAIN's original CLIP-based visual backbone ($\mathcal{V}_{CLIP}$) with DINOtxt~\cite{simeoni2025dinov3} ($\mathcal{V}$).
Second, we restrict the DETR within LAIN to human-object localization only. 
In the original LAIN, DETR is used not only to predict the human and object bounding boxes, $\mathbf{b}_h$ and $\mathbf{b}_o$, but also to extract the corresponding region features, $\mathbf{f}_h$ and $\mathbf{f}_o$, for downstream interaction modeling~\cite{kim2025LAIN}. In our variant, consistent with our proposed \method, DETR is used only to predict $\mathbf{b}_h$ and $\mathbf{b}_o$.
To this end, given $\mathbf{b}_h$ and $\mathbf{b}_o$, we extract patch-level features $\mathbf{F}$ from the DINOtxt visual encoder and select the patches that lie inside the corresponding human and object boxes. We then compute a uniform average pooling over the selected patch features, obtaining one feature vector for the human ($\mathbf{f}_h$) and one for the object ($\mathbf{f}_o$). These pooled DINOtxt features replace the detector-derived region features used in the original LAIN.

\section{Additional experiments}
\label{sec:supp_experiments}
\begin{table*}[!t]
\caption{Ablation of \method on HICO-DET dataset across UO, UC-RF, UC-NF, and UC settings. UV setting is shown in~\cref{tab:main:abl_hicodet_uv} SM stands for soft-masks (\cref{sec:main:method:object_detector_and_soft_mask}), APA for \adaptivepatchmodule (\cref{sec:main:method:adaptive_grounding}), and $\mathcal{I}$ for the \regiontransformermodule (\cref{sec:main:method:interaction_transformer}). Best results in bold.}
\label{tab:supp:abl_hicodet_uo_rf_nf_uc}
\resizebox{\textwidth}{!}{%
\begin{tabular}{@{}ccc|cccc|cccc|cccc|cccc@{}}
\toprule
\multicolumn{3}{c|}{} & \multicolumn{4}{c|}{\textbf{UO}} & \multicolumn{4}{c|}{\textbf{UC-RF}} & \multicolumn{4}{c|}{\textbf{UC-NF}} & \multicolumn{4}{c}{\textbf{UC}} \\
\textbf{SM} & \textbf{APA} & \textbf{$\mathcal{I}$}
& \textbf{Full} & \textbf{Seen} & \textbf{Unseen} & \textbf{HM}
& \textbf{Full} & \textbf{Seen} & \textbf{Unseen} & \textbf{HM}
& \textbf{Full} & \textbf{Seen} & \textbf{Unseen} & \textbf{HM}
& \textbf{Full} & \textbf{Seen} & \textbf{Unseen} & \textbf{HM} \\
\midrule
\xmark & \xmark & \xmark
& 33.00 & 33.40 & 30.98 & 32.19
& 34.02 & 35.72 & 27.27 & 31.50
& 30.56 & 30.54 & 30.65 & 30.60
& 33.70 & 33.51 & 31.17 & 32.34 \\
\cmark & \xmark & \xmark
& 34.11 & 33.56 & 36.88 & 35.22
& 35.13 & 35.36 & 34.20 & 34.88
& 34.70 & 34.90 & 33.87 & 34.38
& 35.42 & 36.16 & 32.68 & 34.42 \\
\cmark & \cmark & \xmark
& 35.97 & 35.47 & 38.48 & 36.98
& 37.06 & 37.55 & 35.09 & 36.32
& 35.40 & 35.71 & 34.15 & 34.93
& 36.06 & 36.79 & 33.16 & 34.98 \\
\cmark & \cmark & \cmark
& \textbf{36.62} & \textbf{35.92} & \textbf{40.13} & \textbf{38.03}
& \textbf{38.39} & \textbf{38.55} & \textbf{37.79} & \textbf{38.17}
& \textbf{36.58} & \textbf{36.30} & \textbf{38.11} & \textbf{37.15}
& \textbf{36.88} & \textbf{38.00} &	\textbf{34.79}	& \textbf{36.10} \\
\bottomrule
\end{tabular}%
}
\end{table*}

\noindent \textbf{Component analysis on UO, UC-RF, UC-NF, UC}.
\cref{tab:supp:abl_hicodet_uo_rf_nf_uc} complements the components analysis of \method (\cref{sec:main_experiments}) to the remaining zero-shot settings on HICO-DET, namely UO, UC-RF, UC-NF, and UC. Consistent with the UV setting in~\cref{tab:main:abl_hicodet_uv}, progressively enabling each component leads to steady improvements, confirming that the proposed modules provide complementary benefits across different zero-shot splits.

In the UO setting, incorporating the soft-mask module (SM) yields a clear improvement over the baseline, increasing Unseen mAP by \textbf{+5.90} and HM by \textbf{+3.03}. Adding \adaptivepatchmodule (APA) provides a further gain of \textbf{+1.60} Unseen mAP and \textbf{+1.76} HM, while the final inclusion of the \regiontransformermodule ($\mathcal{I}$) leads to the best overall performance, reaching \textbf{40.13} Unseen mAP and \textbf{38.03} HM. These results indicate that each component contributes to stronger generalization in the unseen-object setting.

A similar trend is observed in the UC-RF setting. Relative to the baseline, SM improves Unseen mAP by \textbf{+6.93} and HM by \textbf{+3.38}, showing that patch-level grounding substantially benefits compositional generalization. APA further improves Unseen mAP by \textbf{+0.89} and HM by \textbf{+1.44}, and incorporating $\mathcal{I}$ yields the strongest final model, improving Full mAP to \textbf{38.39}, Seen mAP to \textbf{38.55}, Unseen mAP to \textbf{37.79}, and HM to \textbf{38.17}.

In the UC-NF setting, the effect of SM is again substantial, improving Full mAP by \textbf{+4.14}, Seen mAP by \textbf{+4.36}, Unseen mAP by \textbf{+3.22}, and HM by \textbf{+3.78}. APA yields a further but more moderate improvement, increasing Unseen mAP by \textbf{+0.28} and HM by \textbf{+0.55}. Finally, adding $\mathcal{I}$ produces the best overall results, with Full mAP reaching \textbf{36.58}, Unseen mAP \textbf{38.11}, and HM \textbf{37.15}. This suggests that contextual interaction modeling remains important even when strong localized patch grounding is already in place.

For the more challenging UC setting, we observe the same overall behavior. SM improves Full mAP by \textbf{+1.72}, Seen mAP by \textbf{+2.65}, Unseen mAP by \textbf{+1.51}, and HM by \textbf{+2.08} over the baseline. APA provides a further consistent improvement, increasing Seen mAP by \textbf{+0.63}, Unseen mAP by \textbf{+0.48}, and HM by \textbf{+0.56}. Adding $\mathcal{I}$ yields the strongest final model, with further gains of \textbf{+0.82} Full mAP, \textbf{+1.21} Seen mAP, \textbf{+1.63} Unseen mAP, and \textbf{+1.12} HM over the SM+APA variant.

Overall, the results across all zero-shot settings confirm that the three components are complementary. SM consistently improves spatial grounding, APA strengthens part-aware interaction modeling and contributes additional gains in unseen generalization, while the \regiontransformermodule integrates localized interaction cues with broader contextual information to produce the strongest overall performance. \\

\begin{figure}[!t]
    \centering
    \includegraphics[width=\columnwidth]{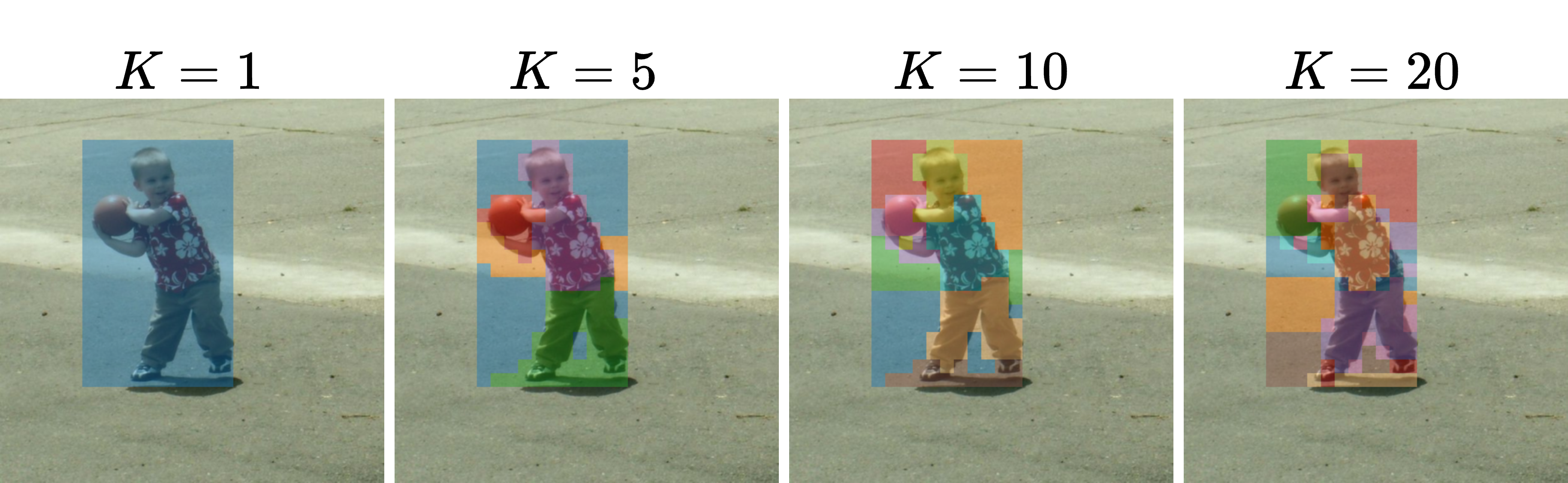}
    \caption{Qualitative results of the unsupervised part decomposition (\cref{sec:main:method:adaptive_grounding}) with $K=\{1, 5, 10, 20\}$.}
    \label{fig:supp:qualitative_k}
\end{figure}

\noindent \textbf{Effectiveness of $\mathbf{K}$}.
\begin{table}[!t]
\caption{Effectiveness of $K$ on HICO-DET under the UV setting. Best results in bold.}
\label{tab:main:abl_hicodet_uv_k}
\resizebox{0.5\columnwidth}{!}{%
\begin{tabular}{@{}lcccc@{}}
\toprule
\textbf{$K$} & \textbf{Full} & \textbf{Seen} & \textbf{Unseen} & \textbf{HM} \\ \midrule
1 & 36.76 & 37.40 & 32.90 & 35.15 \\
5 & 37.37 & 37.90 & 34.18 &	36.04 \\
10 & \textbf{38.10} & \textbf{38.51} & \textbf{35.58} & \textbf{37.04} \\
20 & 37.02	& 37.50	& 34.13 & 35.81 \\
\bottomrule
\end{tabular}%
}
\end{table}

\cref{tab:main:abl_hicodet_uv_k} together with the qualitative analysis in~\cref{fig:supp:qualitative_k} illustrate the effect of varying the number of parts produced by the unsupervised decomposition (\cref{sec:main:method:adaptive_grounding}). 
We observe that $K{=}10$ provides the best trade-off across all metrics. Notably, even with a larger number of parts (e.g., $K{=}20$), the performance remains strong and still surpasses prior state-of-the-art methods (\cref{tab:main:hicodet_zs}), indicating that the gains of \method are not tied to a specific choice of $K$. 

From a modeling perspective, $K$ controls the granularity of the decomposition. As shown in~\cref{fig:supp:qualitative_k}, using too few parts (e.g., $K{=}1$) leads to overly coarse representations that merge interaction-relevant regions, limiting the ability to capture localized cues. Conversely, larger values of $K$ (e.g., $K{=}20$) tend to fragment semantically coherent regions (e.g., splitting head/hair or background regions), which introduces redundancy rather than additional useful interaction evidence. 

Overall, these results suggest that HOI recognition benefits from a moderate level of part granularity, where the decomposition is sufficiently fine to isolate interaction-critical regions while avoiding unnecessary fragmentation. We therefore fix $K{=}10$ in all experiments.

\section{Additional qualitative results}
\label{sec:supp:qualitative}
\begin{figure*}
    \centering
    \includegraphics[height=0.9\textheight]{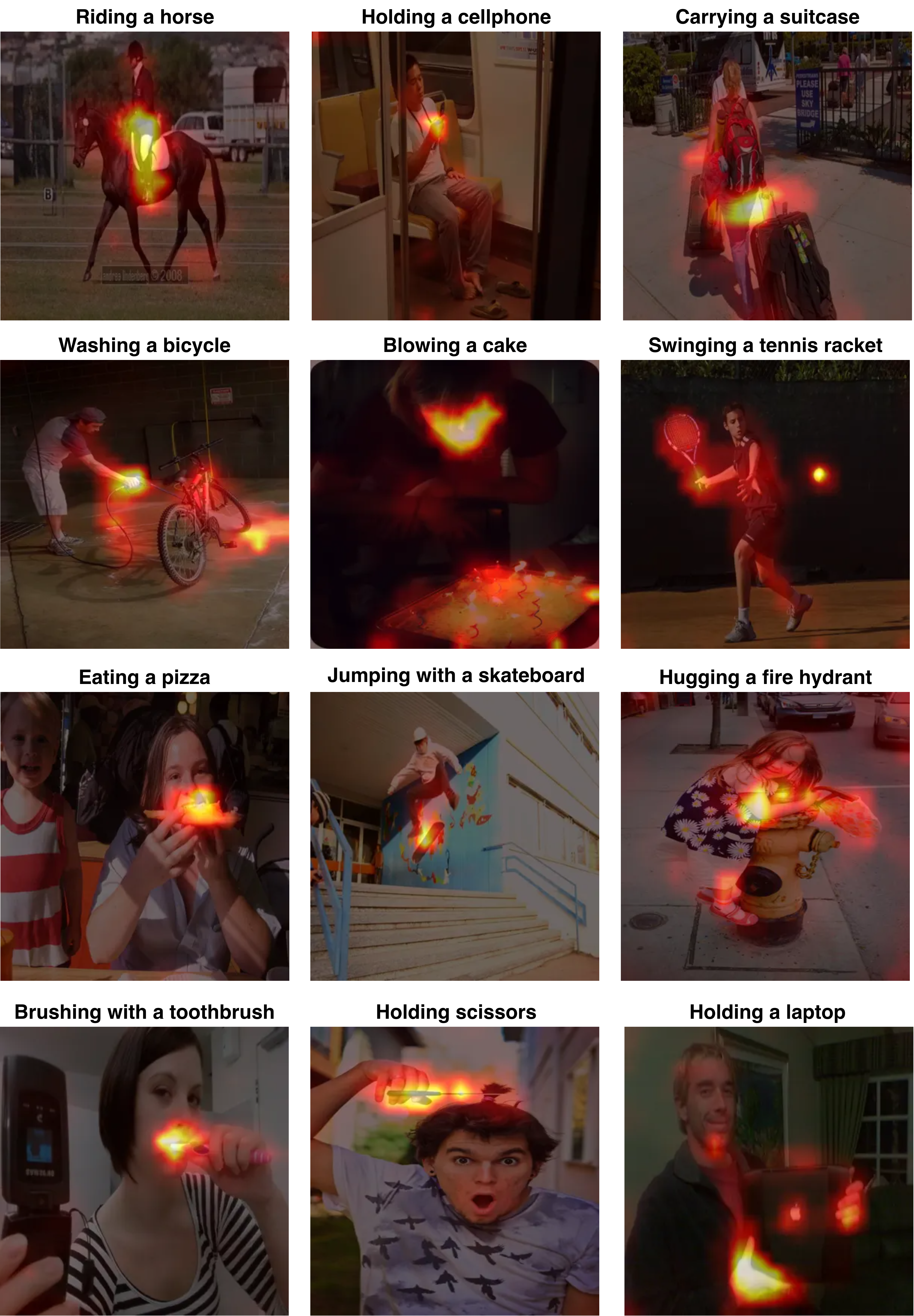}
    \caption{Attention maps of the interaction transformer $\mathcal{I}$ in \method. Warmer colors indicate higher attention to visual patches.}
    \label{fig:supp:qualitatives}
\end{figure*}

\cref{fig:supp:qualitatives} presents additional qualitative examples of the attention maps produced by the interaction transformer $\mathcal{I}$, illustrating how \method localizes interaction cues. Consistent with the findings in~\cref{sec:main:qualitatives}, these examples show that the attention consistently focuses on localized regions that are directly relevant to the predicted interaction, such as the hand and laptop in $\langle$\textit{human}, \textit{holding}, \textit{laptop}$\rangle$, the hand, racket, and ball in $\langle$\textit{human}, \textit{swinging}, \textit{tennis racket}$\rangle$, and the mouth region in $\langle$\textit{human}, \textit{eating}, \textit{pizza}$\rangle$.

\end{document}